\documentclass[10pt,twocolumn,letterpaper]{article}

\usepackage{arxiv}
\usepackage{multirow}
\usepackage{algorithm}
\usepackage{algpseudocode}
\usepackage{colortbl}
\usepackage{array}

\definecolor{linkblue}{rgb}{0.15,0.40,0.68}
\usepackage[breaklinks,colorlinks,allcolors=linkblue]{hyperref}

\definecolor{relaygreen}{RGB}{0,158,115}
\definecolor{relayorange}{RGB}{213,94,0}
\definecolor{ours}{HTML}{EAF1FB}
\definecolor{fam}{HTML}{F2F2F2}

\newcommand{\method}{RELAY}
\newcommand{\methodlong}{Readout-Driven Early Stopping and Visual Relay}
\input{results}

\arxivheader{%
  \includegraphics[height=8.2mm]{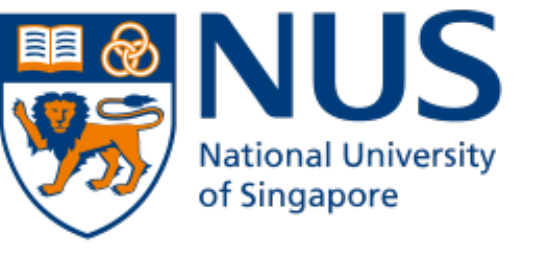}%
  \hfill%
  \includegraphics[height=8.5mm]{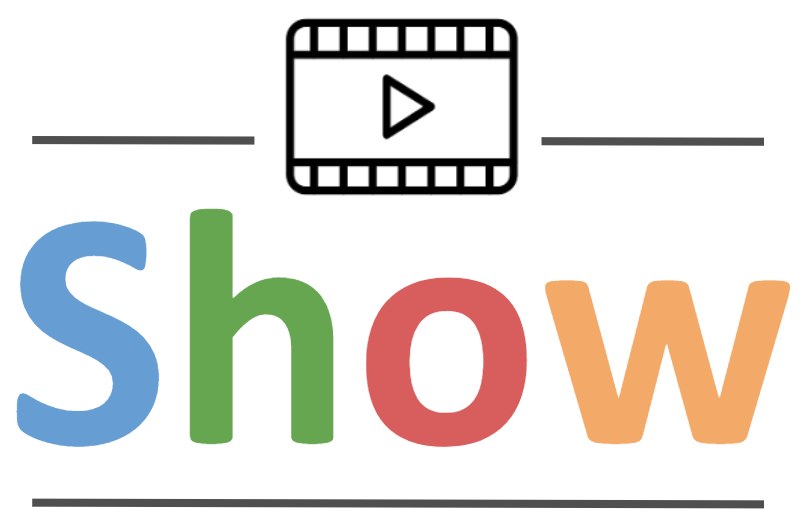}%
}

\title{RELAY: Readout-Driven Early Stopping and Visual Relay\\for Multimodal Multi-Agent Reasoning}

\author{%
  Binjie Zhang$^{1,*}$\quad
  Jiahao Tang$^{2,*}$\quad
  Ling Xing$^{3}$\quad
  Alex Jinpeng Wang$^{2,\dagger}$\quad
  Mike Zheng Shou$^{1,\dagger}$%
}

\affiliation{%
  $^{1}$Show Lab, National University of Singapore\\[1.5pt]
  $^{2}$Central South University \qquad $^{3}$Nanjing University of Science and Technology%
}

\authornote{%
  $^{*}$Equal contribution \qquad $^{\dagger}$Corresponding authors%
}

\begin{document}
\maketitle
\begin{abstract}
Multi-agent collaboration has become an effective test-time paradigm for complex reasoning in vision--language models (VLMs), yet existing methods execute a fixed chain of agents on every input regardless of question difficulty.
On easy multimodal questions, running additional agents causes overthinking and wastes tokens; on hard questions, passing unfinished flawed reasoning text across agents accumulates and propagates early visual errors.
To overcome both limitations, we propose \textbf{\method{}} (\emph{\methodlong{}}), a simple, training-free early-stopping framework for multimodal multi-agent reasoning based on two core designs.
First, we introduce {Readout-Driven Early Stopping}: while the initial agent is reasoning, we periodically read out its tentative answer and stop generation early once the prediction stabilizes with high confidence.
Second, we design {Context-Reset Visual Relay}: when early readouts fail to stabilize with high confidence, we stop the initial agent early and reset the context by discarding its unfinished reasoning text.
We then crop task-relevant image regions and pass only the visual information to a verification agent to re-evaluate the question in a clean context.
Coupling these two designs saves tokens on easy questions and prevents error propagation on hard ones.
Across MMStar, MathVista, and WeMath on Qwen3-VL-8B and 4B, \method{} improves accuracy by \res{h1_pt}\% and \res{m2_h5_pt}\% over \texttt{LatentMAS}, reduces decoding token cost by 24\%--39\%, and accelerates end-to-end inference by $1.37\times$--$1.60\times$, while transferring to text-based reasoning via Stage~1 early stopping without accuracy degradation relative to \texttt{LatentMAS}.
\end{abstract}

\begin{figure*}[t]
  \centering
  \includegraphics[width=0.92\textwidth]{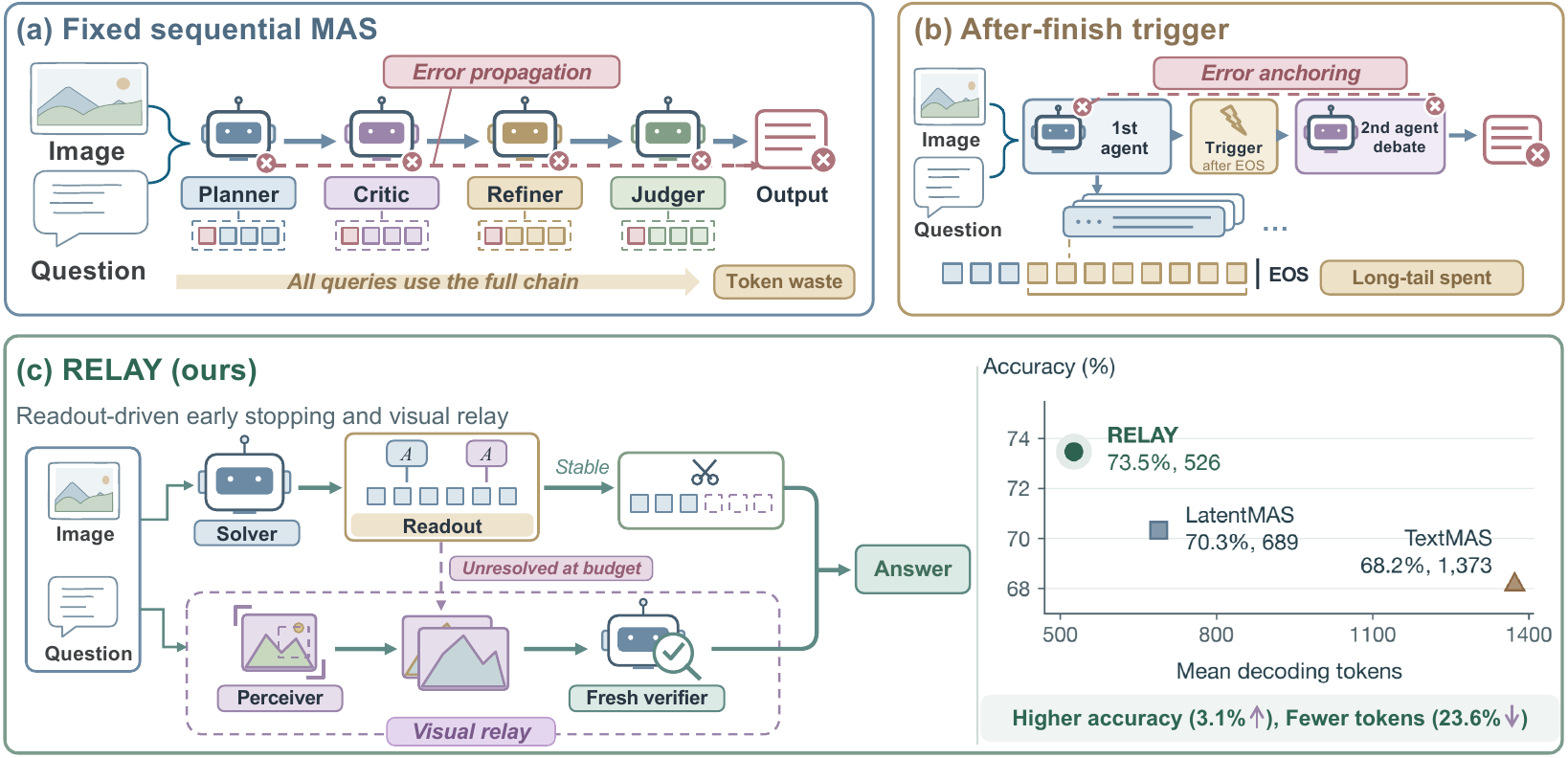}
  \vspace{-3.5mm}
  \caption{\textbf{Comparison of multimodal multi-agent collaboration paradigms.}
    \emph{(a) Fixed Sequential MAS (LatentMAS~\cite{latentmas} / TextMAS~\cite{TextMAS:Zhang2024ChainofAgents}):} Runs all four agents on every input and passes reasoning traces downstream, wasting tokens on easy queries and propagating early errors.
    \emph{(b) After-Finish Trigger (DOWN~\cite{down} / iMAD~\cite{imad}):} Checks uncertainty only after the first agent completes its full response (EOS), failing to save long-tail tokens and anchoring the second agent to initial mistakes.
    \emph{(c) \method{} (Ours):} Reuses a Shared Visual Cache and monitors the Solver with periodic latent readouts to stop early once answers stabilize; when early readouts fail to converge, Context-Reset Visual Relay stops the Solver early, discards the flawed trace, and passes localized Perceiver crops to a fresh Verifier.}
  \label{fig:overview}
  \vspace{-3.5mm}
\end{figure*}

\section{Introduction}
\label{sec:intro}

In large language models (LLMs), test-time multi-agent collaboration is a powerful paradigm for complex reasoning~\cite{camel,du2024debate,liang2024mad,latentmas,tang2026charts,mao2026textground4m,tang2026vectorharness}.
By coordinating multiple models through sequential roles such as planner, critic, refiner, and judger, multi-agent systems (MAS) verify intermediate outputs and improve accuracy without updating model weights~\cite{xu2026vs,kang2026viki}.
However, applying text-based sequential collaboration to multimodal VLMs overlooks how visual perception interacts with multi-step reasoning.

Specifically, when extending existing collaboration paradigms to VLMs (\cref{fig:overview}a, b), we identify two fundamental obstacles:
(1) \emph{Severe token waste and overthinking on easy multimodal questions:} because images already provide dense spatial context, most multimodal questions converge to a correct answer early during generation. Continuing generation or invoking extra agents causes \emph{overthinking}~\cite{s1,han2025tokenbudget,deer} that wastes tokens and can degrade accuracy: \textbf{Fixed Sequential MAS} (\texttt{TextMAS}~\cite{TextMAS:Zhang2024ChainofAgents} or \texttt{LatentMAS}~\cite{latentmas}, \cref{fig:overview}a) runs all agents on every easy input, while \textbf{After-Finish Triggers} (DOWN~\cite{down} and iMAD~\cite{imad}, \cref{fig:overview}b) wait until the full long-tail response finishes.
(2) \emph{Visual error accumulation and cross-agent anchoring:} on hard questions where the initial agent misreads fine-grained visual details, continuing generation rarely self-corrects~\cite{huang2024selfcorrect,laban2026lost}; instead, early errors accumulate and trap the model in a repetitive reasoning loop. Passing this flawed trace downstream anchors subsequent agents to repeat the initial mistake~\cite{smit2024mad,wang2024rethinking,kim2026agentscaling} rather than re-examining the image.

Driven by this analysis, we present \textbf{\method{}} (\emph{\methodlong{}}, \cref{fig:overview}c), a training-free early-stopping framework for multimodal multi-agent reasoning based on two core designs.
First, we introduce \textbf{Readout-Driven Early Stopping}: while the initial agent is reasoning, we periodically probe its tentative answer and stop generation early once the prediction stabilizes with high confidence, avoiding overthinking and redundant agent calls on easy questions.
Second, we design \textbf{Context-Reset Visual Relay}: when early readouts fail to converge on hard questions, we stop the initial agent early, discard its unfinished reasoning text, and pass localized task-relevant image crops to a verification agent in a clean context.
Because resetting the context decouples verification from the first agent's stopping step, we further introduce \textbf{Offline Trace-Replay Policy Search (TRPS)} to replay cached development traces and select early-stopping thresholds in seconds without extra VLM inference.

We evaluate \method{} across standard multimodal benchmarks (MMStar~\cite{mmstar}, MathVista~\cite{mathvista}, WeMath~\cite{wemath}), high-resolution visual datasets~\cite{vstar,hrbench,mmerw}, and text benchmarks (ARC~\cite{arc}, GSM8K~\cite{gsm8k}) on Qwen3-VL-8B, Qwen3-VL-4B~\cite{qwen3vl}, InternVL3.5-8B-HF~\cite{wang2025internvl3}, and Qwen3-8B~\cite{yang2025qwen3}, spanning both VLMs and LLMs.
\method{} improves accuracy by \res{h1_pt}\% on 8B and \res{m2_h5_pt}\% on 4B over \texttt{LatentMAS}, cuts decoding tokens by 24\%--39\%, accelerates end-to-end B200 inference by $1.37\times$--$1.60\times$, and transfers directly to text-based reasoning via Stage~1 early stopping (+\res{h3_pt}\% accuracy with 20.2\% fewer tokens over \texttt{LatentMAS}).

In summary, our main contributions are:
\begin{itemize}[leftmargin=*,itemsep=1.5pt,topsep=1.5pt]
  \item \textbf{Multimodal Multi-Agent Early-Stopping Paradigm:} We identify severe token waste from overthinking on easy queries and visual error accumulation on hard queries in existing multimodal MAS, and propose \textbf{\method{}}, a training-free early-stopping framework for multimodal multi-agent reasoning.
  \item \textbf{Context-Reset Visual Relay and Offline TRPS:} To prevent visual error propagation from flawed reasoning traces, we introduce Context-Reset Visual Relay with localized image crops; this trace-free design further enables Offline TRPS to replay cached traces and lock early-stopping thresholds without extra VLM inference.
  \item \textbf{Effectiveness, Efficiency, and Transfer:} \method{} improves accuracy over fixed sequential MAS while substantially reducing decoding token cost and wall-clock latency across model scales, and transfers directly to text-only reasoning via Stage~1 early stopping.
\end{itemize}

\section{Related Work}
\label{sec:related}

\noindent\textbf{Multi-agent collaboration and adaptive triggering.}
Existing test-time MAS generally communicate through two paradigms that differ in exchanged representations: \emph{text-based collaboration}, where agents exchange generated text via debate, role-playing, or sequential workflows~\cite{camel,du2024debate,liang2024mad,autogen,vif}, and \emph{KV-cache or latent collaboration}, such as LatentMAS~\cite{latentmas} and related continuous communication methods~\cite{cipher,actcomm,c2c,visionwormhole,l2vmas} that pass internal hidden states or KV caches across agents.
When applied to multimodal reasoning, both paradigms operate as \textbf{Fixed Sequential MAS} (\cref{fig:overview}a): they execute all agents on every question and pass the initial agent's reasoning trace downstream, wasting tokens on easy visual questions and anchoring subsequent agents to earlier mistakes~\cite{smit2024mad,wang2024rethinking,huang2024selfcorrect,laban2026lost,kim2026agentscaling}.
To reduce overhead, adaptive methods prune agent graphs~\cite{dylan,agentprune,agentdropout}, train routing modules~\cite{ghosh2026grade,ghasemabadi2026multihead}, or apply training-free \textbf{After-Finish Triggers} such as DOWN~\cite{down}, iMAD~\cite{imad}, and perceptual-disagreement gates~\cite{lase,gamagent,eagle} (\cref{fig:overview}b), which evaluate uncertainty only after the first agent completes its full response.
Unlike both paradigms, \method{} (\cref{fig:overview}c) performs early stopping via periodic readouts once answers stabilize and resets the context when reasoning fails to converge, sharing only visual evidence~\cite{vllm,sglang,kvcomm} without inheriting flawed reasoning traces.

\vspace{8pt}
\noindent\textbf{Visual search and region zooming.}
Downsampling high-resolution images degrades fine-grained visual recognition as target objects shrink~\cite{hrbench,vicrop}.
While training-based approaches teach models to localize or zoom via supervised fine-tuning~\cite{vstar,visualcot} or reinforcement learning~\cite{openai2025thinking,deepeyes,chainoffocus,minio3}, training-free methods zoom over sub-image trees~\cite{zoomeye,dyfo}, attention maps~\cite{vicrop}, or patch descriptions~\cite{hrbench}, typically appending zoomed views to the ongoing trajectory.
In contrast, \method{} triggers localized cropping via early stopping only when the initial agent fails to converge and relays the crops into a reset context free of prior reasoning errors.

\vspace{8pt}
\noindent\textbf{Early stopping and offline policy search.}
Test-time compute scaling adapts generation effort to problem difficulty~\cite{calm,snell2025scaling,s1,han2025tokenbudget}.
Early-stopping self-consistency~\cite{selfconsistency,esc,adaptiveconsistency} halts multi-chain sampling when answers agree, while single-chain methods exit on trial-answer confidence (DEER~\cite{deer}), answer stability (Certaindex~\cite{certaindex}), trace confidence filtering (DeepConf~\cite{deepconf}), probability jumps~\cite{tikhonov2026confidenceleaps}, hesitation penalties~\cite{lotfi2026penalty}, or model cascades~\cite{motcascade}, subject to readout calibration properties~\cite{zheng2024selectors,li2026readout}.
However, existing early-stopping methods, designed for text-only reasoning, mainly regulate when textual generation terminates; they cannot resolve fine-grained visual perception bottlenecks when an agent gets stuck or prevent error propagation across multiple agents.
\method{} bridges this gap by coupling readout-driven early stopping with context-reset visual relay and offline trace-replay policy search~\cite{dspy,mipro,gepa}.

\section{Methodology}
\label{sec:method}

\method{} is a simple, training-free early-stopping framework for multimodal multi-agent reasoning, built on two online designs, {Readout-Driven Early Stopping (Stage~1)} and {Context-Reset Visual Relay (Stage~2)}, coupled with training-free {Offline Trace-Replay Policy Search (TRPS)} (\cref{fig:overview}c, \cref{fig:method_flow}).
We first present our motivating analysis and problem formulation (\cref{sec:diagnosis}), and then describe our online early-stopping and visual relay framework together with offline policy search (\cref{sec:readout}).

\subsection{Motivation and Problem Formulation}
\label{sec:diagnosis}

\begin{figure}[t]
  \centering
  \includegraphics[width=\linewidth]{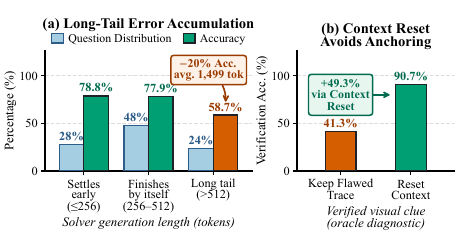}
  \vspace{-7.5mm}
  \caption{\textbf{Empirical motivation on Qwen3-VL-8B.}
    \emph{(a) Long-Tail Error Accumulation:} Bars show question distribution and the Solver's accuracy on the 921-question test set. Over 75\% of questions finish within 512 tokens at $\sim$78\% accuracy, whereas unresolved long-tail cases ($>512$ tokens, avg.\ 1{,}499) drop by $\sim$20\% in accuracy.
    \emph{(b) Context Reset Avoids Anchoring:} In an oracle diagnostic on 12 development questions (25 states, 3 samples per state) given the same verified visual clue, the next agent reaches 41.3\% accuracy when keeping the flawed trace, compared with 90.7\% (+49.3\%) after resetting the prior reasoning context.}
  \label{fig:motivation_stats}
  \vspace{-3.5mm}
\end{figure}

\paragraph{Motivating observations.}
Why do existing multi-agent paradigms struggle on visual reasoning?
Our empirical analysis on Qwen3-VL-8B reveals a sharp divergence between fast-converging trajectories and unresolved long-tail trajectories (\cref{fig:motivation_stats}).
When global visual features suffice, most visual questions converge early, settling within 256 tokens or finishing by 512; in contrast, when fine-grained visual details are ambiguous, generation drags into a long tail ($>$512 tokens) that consumes several times more tokens while accuracy drops sharply (\cref{fig:motivation_stats}a).
Furthermore, in an oracle diagnostic on 12 selected development questions where the initial agent misreads visual details, appending a verified visual clue to its flawed reasoning trace yields 41.3\% accuracy, whereas resetting the context lets the agent use the same clue and reach 90.7\% (\cref{fig:motivation_stats}b).
In summary, existing multimodal MAS suffer from two fundamental bottlenecks: \emph{(1) severe token waste and overthinking on easy visual questions} when running fixed chains or waiting for full completion (\cref{fig:overview}a, b), and \emph{(2) visual error accumulation and cross-agent anchoring} when flawed long-tail traces are passed across agents.

\paragraph{Problem formulation.}
These observations motivate a two-stage early-stopping and visual relay framework that stops generation early once the answer stabilizes with high confidence on easy inputs, and stops the Solver early to reset the context with localized visual evidence when early readouts fail to converge (\cref{fig:overview}c, \cref{fig:method_flow}).
Given an input image $I$ and question $q$, \method{} coordinates three roles from a single frozen VLM: an initial \textbf{Solver}, a visual localization \textbf{Perceiver}, and a trace-free \textbf{Verifier}.
Our goal is to minimize decoding token cost while preserving or improving reasoning accuracy across diverse tasks.

\subsection{RELAY}
\label{sec:readout}

\paragraph{Shared visual cache and context reset.}
To avoid redundant image encoding while preventing cross-agent error propagation, we separate visual feature sharing from reasoning trace transfer (\cref{fig:overview}c).
We encode the system prompt and input image $I$ once into a \textbf{Shared Visual Cache} $R$.
Unlike fixed sequential MAS that append the initial agent's reasoning trace $\tau_S$ to downstream agents, \method{} resets the context when handing over to the Verifier:
\begin{equation}
  H_S = R \oplus q, \quad
  H_V = R \oplus C_{\mathrm{crop}} \oplus q \quad (\tau_S \cap H_V = \varnothing),
  \label{eq:substrate}
\end{equation}
where $H_S$ and $H_V$ denote the input contexts of the Solver and the Verifier, $\tau_S$ denotes the Solver's generated reasoning trace, and $C_{\mathrm{crop}}$ denotes the localized image crops from the Perceiver.
By resetting the context in $H_V$ ($\tau_S \cap H_V = \varnothing$) instead of appending the Solver's unfinished reasoning text, this design reuses the encoded visual features $R$ while eliminating error anchoring from flawed reasoning traces.

\begin{figure}[t]
  \centering
  \includegraphics[width=\linewidth]{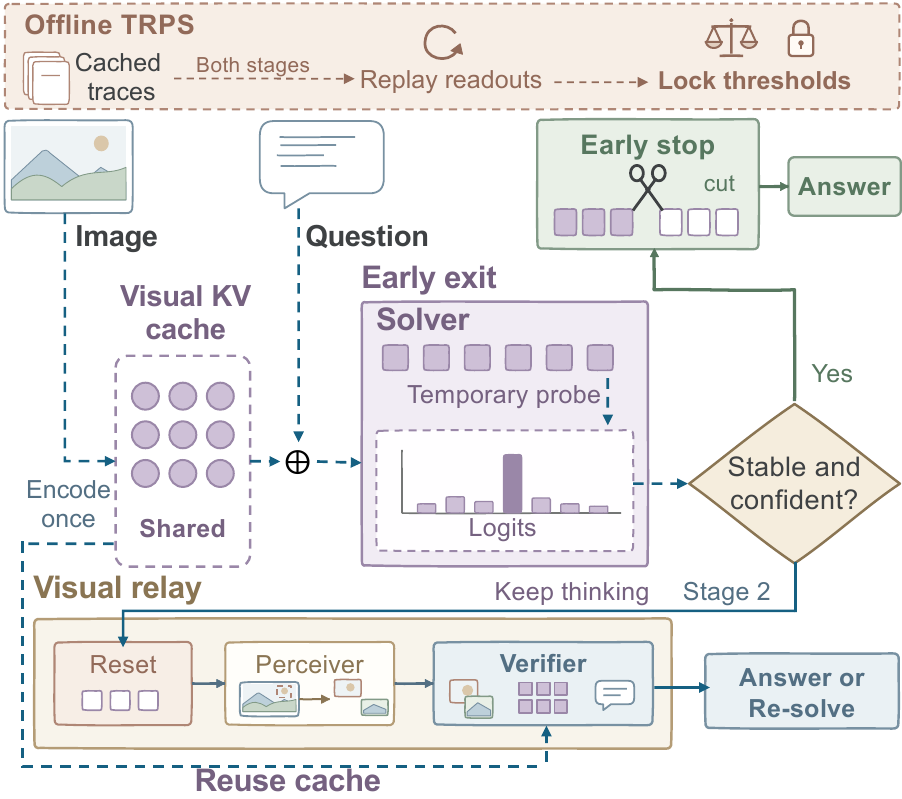}
  \vspace{-4.5mm}
  \caption{\textbf{Design overview of \method{}.} Online inference monitors the Solver via periodic latent readouts to stop generation early once the answer stabilizes with high confidence (Stage~1), and stops unconverged runs early to reset the context and relay localized Perceiver crops to a trace-free Verifier (Stage~2). Because readout probing and context reset decouple verification from the Solver's stopping step, Offline TRPS can replay cached traces to lock stopping thresholds without extra VLM inference.}
  \label{fig:method_flow}
  \vspace{-3.5mm}
\end{figure}

\paragraph{Latent Answer Readout Probe.}
To monitor an agent's predicted answer during generation without disrupting its reasoning flow, we introduce a lightweight \textbf{Latent Answer Readout Probe} (\emph{latent readout}, \cref{fig:method_flow}).
After the agent generates a certain number of tokens, we branch a temporary copy of its current KV cache $C_t$, append a short answer-extraction prompt, and run a single forward pass to read out its tentative answer $\hat{a}_t$ and confidence $c_t$:
\begin{equation}
  (\hat{a}_t, c_t) = \operatorname{Readout}(C_t).
  \label{eq:readout}
\end{equation}
The temporary KV branch is discarded immediately after each readout so that the agent's ongoing generation remains unchanged (see appendix for details).

\paragraph{Two-stage routing by readout-estimated difficulty.}
We use the Latent Answer Readout Probe to estimate difficulty during generation and route queries through two stages (\cref{fig:method_flow}).
VLMs solving easy visual questions are prone to \emph{overthinking}, where continuing generation after the answer converges wastes tokens and can degrade accuracy; conversely, when solving hard visual questions, early perceptual errors accumulate and trap the model in a repetitive reasoning loop.
Based on whether early readouts stabilize with high confidence, \method{} triggers \textbf{Stage~1 (Readout-Driven Early Stopping)} once the answer converges early on easy questions and \textbf{Stage~2 (Context-Reset Visual Relay)} when early readouts fail to converge on hard questions.

\paragraph{Stage 1: Early stopping on stabilized answers.}
Stage~1 stops generation early on easy questions where global visual features already suffice (\cref{fig:method_flow}).
As the Solver reasons (exiting naturally if it finishes early), \method{} periodically performs latent readouts (\cref{eq:readout}) to track its tentative answer and confidence.
Once consecutive readouts agree on the same answer and the confidence exceeds a stopping threshold, the Solver halts immediately and returns that answer.
Combining high confidence filtering with checkpoint agreement prevents premature exits on uncertain guesses while cutting redundant tail generation.

\begin{table*}[t]
\centering
\footnotesize
\setlength{\tabcolsep}{4pt}
\renewcommand{\arraystretch}{1.02}
\resizebox{\textwidth}{!}{%
\begin{tabular}{l c cccc c c}
\toprule
\multirow{2}{*}{\textbf{Method}}
& \multirow{2}{*}{\textbf{Agents}}
& \multicolumn{4}{c}{\textbf{Accuracy (\%)}}
& \multirow{2}{*}{\textbf{Avg.\ Tokens}}
& \multirow{2}{*}{\textbf{Token Change vs.\ 4-Agent MAS}} \\
\cmidrule(lr){3-6}
& & \textbf{MMStar} & \textbf{MathVista} & \textbf{WeMath} & \textbf{All} & & \\
\midrule
\rowcolor{fam}[\tabcolsep][\tabcolsep]
\multicolumn{8}{c}{\textit{Qwen3-VL-8B (3 seeds, 921 test questions)}} \\
Single VLM (Default) & 1 & \res{acc_single_mmstar} & \res{acc_single_mathvista} & \res{acc_single_wemath} & \res{acc_single} & \res{out_single} & \textcolor{relaygreen}{$\downarrow$ 23.2\%} ($0.77\times$) \\
Self-Consistency (3 Chains) & 3 & 68.1 & 84.1 & 76.0 & \res{acc_sc3} & \res{out_sc3} & \textcolor{relayorange}{$\uparrow$ 159.5\%} ($2.60\times$) \\
\addlinespace[2pt]
TextMAS (4-Agent) & 4 & \res{acc_tm_mmstar} & \res{acc_tm_mathvista} & \res{acc_tm_wemath} & \res{acc_tm} & \res{out_tm} & \textcolor{relayorange}{$\uparrow$ 99.3\%} ($1.99\times$) \\
LatentMAS (2-Agent, Planner--Judger) & 2 & \res{acc_pj_mmstar} & \res{acc_pj_mathvista} & \res{acc_pj_wemath} & \res{acc_pj} & \res{out_pj} & \textcolor{relayorange}{$\uparrow$ 2.4\%} ($1.02\times$) \\
LatentMAS (4-Agent, Default) & 4 & \res{acc_lmas_mmstar} & \res{acc_lmas_mathvista} & \res{acc_lmas_wemath} & \res{acc_lmas} & \res{out_lmas} & Baseline ($1.00\times$) \\
\addlinespace[2pt]
\rowcolor{ours}[\tabcolsep][\tabcolsep]
\method{} (w/o TRPS, Heuristic) & 1--3 & 68.5 & 82.9 & 72.2 & 73.1 & 641 & \textcolor{relaygreen}{$\downarrow$ 7.0\%} ($0.93\times$) \\
\rowcolor{ours}[\tabcolsep][\tabcolsep]
\textbf{\method{} (Ours, w/ TRPS)} & \textbf{1--3} & \textbf{\res{acc_ours_mmstar}} & \textbf{\res{acc_ours_mathvista}} & \textbf{\res{acc_ours_wemath}} & \textbf{\res{acc_ours}} & \textbf{\res{out_ours}} & \textcolor{relaygreen}{\textbf{$\downarrow$ 23.6\%}} (\textbf{0.76$\times$}) \\
\midrule
\rowcolor{fam}[\tabcolsep][\tabcolsep]
\multicolumn{8}{c}{\textit{Qwen3-VL-4B (3 seeds, 921 test questions)}} \\
Single VLM (Default) & 1 & \res{m2_acc_single_mmstar} & \res{m2_acc_single_mathvista} & \res{m2_acc_single_wemath} & \res{m2_acc_single} & \res{m2_out_single} & \textcolor{relaygreen}{$\downarrow$ 19.9\%} ($0.80\times$) \\
Self-Consistency (3 Chains) & 3 & 65.5 & 84.6 & 77.9 & \res{m2_acc_sc3} & \res{m2_out_sc3} & \textcolor{relayorange}{$\uparrow$ 98.7\%} ($1.99\times$) \\
\addlinespace[2pt]
TextMAS (4-Agent) & 4 & 61.2 & 79.8 & 67.3 & 67.7 & 1{,}292 & \textcolor{relayorange}{$\uparrow$ 78.6\%} ($1.79\times$) \\
LatentMAS (2-Agent, Planner--Judger) & 2 & \res{m2_acc_pj_mmstar} & \res{m2_acc_pj_mathvista} & \res{m2_acc_pj_wemath} & \res{m2_acc_pj} & \res{m2_out_pj} & \textcolor{relaygreen}{$\downarrow$ 0.1\%} ($1.00\times$) \\
LatentMAS (4-Agent, Default) & 4 & \res{m2_acc_lmas_mmstar} & \res{m2_acc_lmas_mathvista} & \res{m2_acc_lmas_wemath} & \res{m2_acc_lmas} & \res{m2_out_lmas} & Baseline ($1.00\times$) \\
\addlinespace[2pt]
\rowcolor{ours}[\tabcolsep][\tabcolsep]
\method{} (w/o TRPS, Heuristic) & 1--3 & 63.7 & 81.1 & 74.0 & 71.6 & 513 & \textcolor{relaygreen}{$\downarrow$ 29.1\%} ($0.71\times$) \\
\rowcolor{ours}[\tabcolsep][\tabcolsep]
\textbf{\method{} (Ours, w/ TRPS)} & \textbf{1--3} & \textbf{\res{m2_acc_ours_mmstar}} & \textbf{\res{m2_acc_ours_mathvista}} & \textbf{\res{m2_acc_ours_wemath}} & \textbf{\res{m2_acc_ours}} & \textbf{\res{m2_out_ours}} & \textcolor{relaygreen}{\textbf{$\downarrow$ 39.1\%}} (\textbf{0.61$\times$}) \\
\midrule
\rowcolor{fam}[\tabcolsep][\tabcolsep]
\multicolumn{8}{c}{\textit{InternVL3.5-8B-HF (1 paired seed, 921 test questions)}} \\
Single VLM (Default) & 1 & 65.5 & 84.1 & 67.8 & 70.5 & 194 & \textcolor{relaygreen}{$\downarrow$ 94.2\%} ($0.06\times$) \\
TextMAS (4-Agent) & 4 & 62.4 & 82.1 & 68.9 & 69.3 & 3{,}376 & Baseline ($1.00\times$) \\
\addlinespace[2pt]
\rowcolor{ours}[\tabcolsep][\tabcolsep]
\method{} (w/o TRPS, Zero-Shot) & 1--3 & 68.1 & 85.1 & 68.3 & 71.9 & 827 & \textcolor{relaygreen}{$\downarrow$ 75.5\%} ($0.25\times$) \\
\rowcolor{ours}[\tabcolsep][\tabcolsep]
\textbf{\method{} (Ours, w/ TRPS)} & \textbf{1--3} & \textbf{69.8} & \textbf{85.6} & \textbf{67.8} & \textbf{72.4} & \textbf{740} & \textcolor{relaygreen}{\textbf{$\downarrow$ 78.1\%}} (\textbf{0.22$\times$}) \\
\bottomrule
\end{tabular}%
}
\vspace{-1.5mm}
\caption{\textbf{Main multimodal reasoning results} across MMStar~\protect\cite{mmstar}, MathVista~\protect\cite{mathvista}, and WeMath~\protect\cite{wemath}.
Gray bands distinguish backbones; blue bands highlight \method{} variants, with bold denoting the full TRPS-calibrated \method{} policy.
Token changes (\textcolor{relaygreen}{$\downarrow$} reduction, \textcolor{relayorange}{$\uparrow$} increase) are computed from full-precision means relative to the 4-agent sequential baseline for each backbone.
\method{} improves accuracy over 4-agent \texttt{LatentMAS} and \texttt{TextMAS} with substantially fewer tokens, whereas Self-Consistency (3 Chains) attains higher accuracy at $2.0\times$--$2.6\times$ token cost.
See appendix for InternVL3.5-8B-HF cap-readout controls and paired intervals.}
\label{tab:main}
\vspace{-3mm}
\end{table*}

\paragraph{Stage 2: Context-reset visual relay.}
When coarse-grained global visual tokens are insufficient for accurate reasoning, early readouts from the Solver fail to stabilize with high confidence (\cref{sec:diagnosis}).
This non-convergence typically reflects missing fine-grained visual details rather than insufficient text steps; continuing from the original trace without resetting only accumulates early errors and lowers accuracy.

Therefore, Stage~2 stops the Solver early once its readouts fail to converge, records its tentative answer, and \emph{resets the context} by discarding the unfinished reasoning trace (\cref{fig:method_flow}, Stage~2).
The Perceiver then localizes and zooms in on task-relevant image regions, passing the cropped regions with the shared visual cache and question to the Verifier in a clean context (\cref{eq:substrate}).
On this clean visual context, the Verifier first performs a one-step latent readout (\emph{Quick Check}, \cref{eq:readout}): if this readout confirms the Solver's tentative answer with high confidence, \method{} outputs it immediately; otherwise, the Verifier solves the question from scratch on the clean visual context, recovering accuracy on hard visual cases while avoiding long-tail token waste.

\label{sec:solver}
\paragraph{Why offline policy search is needed and feasible.}
While Stage~1 and Stage~2 govern online execution, the overall trade-off between accuracy and token cost depends on the early-stopping token steps and confidence thresholds across both stages.
Setting these thresholds manually by heuristics is suboptimal, whereas searching them online in standard sequential MAS requires re-running multi-agent VLM inference for every candidate configuration because changing when the first agent stops alters the text passed downstream.

\paragraph{Training-free threshold selection via trace replay.}
In \method{}, our two core designs make offline search tractable (\cref{fig:method_flow}, top): because Stage~1's latent readout does not alter ongoing generation and Stage~2's Visual Relay resets the context, the Perceiver crops and Verifier outputs depend only on the shared visual cache and question, remaining invariant to the Solver's stopping step.
We therefore cache the Solver trajectory and clean-context Visual Relay outputs once on a development set, and run \textbf{Offline Trace-Replay Policy Search (TRPS)} to replay candidate configurations under a token-cost constraint without weight updates or extra VLM rollouts, locking a single frozen policy for test evaluation (see appendix for details).

\section{Experiments}
\label{sec:exp}

We evaluate \method{} across multimodal, high-resolution visual, and text-only reasoning benchmarks under multiple model scales (\cref{sec:setup,sec:main,sec:text}).
We then ablate the Stage~1 stopping signals, Stage~2 handover timing, and Offline TRPS policy search (\cref{sec:ablation}), and analyze the Stage~2 context-reset visual relay and execution paths (\cref{sec:stops}).

\subsection{Experimental Setup}
\label{sec:setup}

\begin{figure*}[t]
  \centering
  \includegraphics[width=\textwidth]{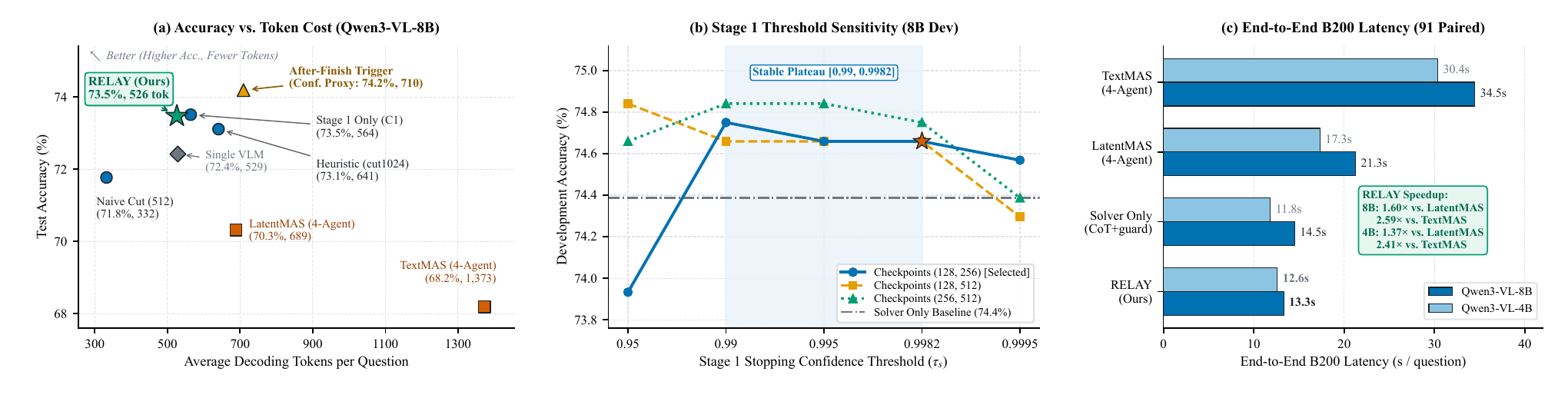}
  \vspace{-4.5mm}
  \caption{\textbf{Accuracy--cost trade-off, Stage 1 threshold stability, and B200 GPU inference latency.}
    \emph{(a) Accuracy vs.\ token cost (Qwen3-VL-8B, 921 test):} \method{} reduces decoding token cost over fixed sequential MAS and after-finish confidence triggers while avoiding the accuracy loss of naive truncation.
    \emph{(b) Stage 1 threshold sensitivity (Qwen3-VL-8B dev, $N{=}367$):} Replayed across checkpoint pairs $(128, 256)$, $(128, 512)$, and $(256, 512)$, high confidence thresholds $[0.99, 0.9982]$ form a stable accuracy plateau above the unstopped Solver baseline.
    \emph{(c) End-to-end B200 latency (91 paired items):} \method{} runs $1.60\times$ faster than 4-agent LatentMAS and $2.59\times$ faster than 4-agent TextMAS on 8B ($1.37\times$ and $2.41\times$ on 4B).}
  \label{fig:exp_summary}
  \vspace{-3mm}
\end{figure*}

\paragraph{Datasets and models.}
We evaluate multimodal reasoning on \res{n_confirm} frozen test questions across MMStar~\cite{mmstar} (\res{n_mmstar}), MathVista~\cite{mathvista} (\res{n_mathvista}), and WeMath~\cite{wemath} (\res{n_wemath}), plus three high-resolution development benchmarks: V*Bench~\cite{vstar}, HR-Bench~\cite{hrbench}, and MME-RealWorld-Lite~\cite{mmerw}.
All agent roles (Solver, Perceiver, Verifier) share one frozen VLM in bf16: Qwen3-VL-8B-Instruct~\cite{qwen3vl} (primary, 3 seeds), Qwen3-VL-4B-Instruct~\cite{qwen3vl} (zero-shot scale transfer, 3 seeds), and InternVL3.5-8B-HF~\cite{wang2025internvl3} (visual-family transfer with a Qwen3 text decoder, 1 paired seed).
For text-only reasoning, we evaluate 900 test problems (300 each from ARC-Easy, ARC-Challenge~\cite{arc}, and GSM8K~\cite{gsm8k}) on Qwen3-8B~\cite{yang2025qwen3}.

\paragraph{Implementation and locked parameters.}
In \method{}, the Solver and Verifier decode up to 2{,}048 tokens (while default Single VLM, TextMAS, and LatentMAS multimodal baselines retain their original 4{,}096-token cap; text evaluation uses 2{,}048 tokens across all methods).
Latent readouts (\cref{eq:readout}) use ``\texttt{Therefore, the final answer is \textbackslash boxed\{}'' (10 tokens on Qwen3-VL) for the Solver and ``\texttt{The final answer is \textbackslash boxed\{}'' (7 tokens) for Quick Check.
In Stage~2, the Perceiver predicts up to two bounding boxes ($\le 128$ tokens); parsed boxes with union area $\le 50\%$ of the image are expanded $3\times$ and resized toward $[448, 1280]$\,px subject to a $2\times$ upscaling cap and visual token budget.
Offline TRPS on the 367-item Qwen3-VL-8B development set locks the Stage~1 readout steps at $\{128, 256\}$ tokens with confidence threshold $\tau_s = 0.9982$, and triggers Stage~2 visual relay at $512$ tokens with Quick Check threshold $\tau_g = 0.9$ across benchmarks (see appendix for threshold ablations and InternVL3.5-8B-HF calibration).

\paragraph{Baselines and metrics.}
We compare three baseline groups:
(1) \textbf{Single-agent and ensemble:} Single VLM and Self-Consistency~\cite{selfconsistency} (majority voting over 3 frozen CoT chains);
(2) \textbf{Fixed sequential MAS:} 4-agent TextMAS~\cite{TextMAS:Zhang2024ChainofAgents} and LatentMAS~\cite{latentmas}, evaluated as both a 2-agent Planner--Judger pipeline and a 4-agent Planner--Critic--Refiner--Judger pipeline;
(3) \textbf{Triggered and ablation variants:} a completion-confidence After-Finish Trigger proxy (using the same Stage~1 Solver and visual relay, inspired by~\cite{down,imad}), an Early-Stop Trigger (at matched call rate), un-tuned \method{} (1{,}024-token structural heuristic on Qwen3-VL; zero-shot transfer on InternVL3.5-8B-HF), and full-length Solver Only (\texttt{CoT+guard}).
We report accuracy (\%), mean decoding tokens, relative token cost, and paired NVIDIA B200 latency (s).

\subsection{Main Multimodal Results and Efficiency}
\label{sec:main}

\paragraph{\method{} outperforms fixed sequential MAS in both accuracy and token efficiency.}
On Qwen3-VL-8B (\cref{tab:main}, \cref{fig:exp_summary}a), fixed 4-agent pipelines (TextMAS and LatentMAS) consume more decoding tokens than a single VLM yet achieve lower accuracy, and increasing latent depth to 40 or 80 steps inflates compute without meaningful accuracy gain (see appendix).
This degradation occurs because unconditional agent chains overthink easy visual questions while propagating early perceptual mistakes across agents on hard ones.
In contrast, \method{} reaches \textbf{\res{acc_ours}\%} accuracy (+\res{h1_pt}\% over 4-agent LatentMAS, 95\% image-cluster CI $[+1.4, +4.9]$) while cutting decoding token cost by \textbf{23.6\%} ($0.76\times$, CI $[0.71, 0.82]$).
Compared with Single VLM, \method{} attains $+1.0\%$ higher point accuracy (CI $[-0.8, +3.0]$) at matched token cost ($526$ vs.\ $529$ tokens), while Self-Consistency (3 Chains) reaches $74.7\%$ at $3.4\times$ the tokens of \method{}.

\paragraph{\method{} transfers across model scales and visual model families.}
Transferring the 8B-selected policy directly to Qwen3-VL-4B (\cref{tab:main}), \method{} improves accuracy by \textbf{+\res{m2_h5_pt}\%} (CI $[+1.7, +5.0]$) over 4-agent LatentMAS while cutting decoding tokens by \textbf{39.1\%} ($0.61\times$, CI $[0.57, 0.65]$), and outperforms 4-agent TextMAS by $+5.0\%$ at $0.34\times$ cost.
Smaller VLMs are more vulnerable to long-chain drift and cross-agent anchoring, yielding larger efficiency and accuracy gains at the 4B scale.

On InternVL3.5-8B-HF (visual-family transfer with a Qwen3 text decoder, 1 paired seed; \cref{tab:main}), dev-calibrated \method{} ($\tau_s{=}0.99, \tau_g{=}1.0$) improves accuracy over default 4-agent TextMAS by $+3.1\%$ while cutting token cost by \textbf{78.1\%} ($0.22\times$).
Two boundaries clarify this transfer (full 6-row breakdown and CIs in appendix): (1) when TextMAS is upgraded with an end-of-cap answer readout, accuracy is comparable ($72.4\%$ vs.\ $73.1\%$, CI $[-3.2, +2.0]$) while \method{} retains the $78.1\%$ token savings, and relative to Single VLM ($194$ tokens), \method{} uses $3.81\times$ tokens ($+2.0\%$, CI $[-0.8, +4.6]$); (2) without prompt adaptation, the Perceiver fails to emit a valid JSON box on $68.8\%$ of questions (primarily from verbose text hitting the 128-token cap), causing $81.8\%$ of Stage~2 handovers to fall back to global-image re-solving.

\begin{table}[t]
  \centering
  \footnotesize
  \setlength{\tabcolsep}{2.6pt}
  \renewcommand{\arraystretch}{1.0}
  \resizebox{\linewidth}{!}{%
  \begin{tabular}{@{}lcccccc@{}}
    \toprule
    Method (Qwen3-8B) & ARC-E & ARC-C & GSM8K & All & Tokens & Token Change vs.\ LatentMAS \\
    \midrule
    Single LLM (Default) & \res{t_single_arce} & \res{t_single_arcc} & \res{t_single_gsm} & \res{t_single} & \res{t_out_single} & \textcolor{relayorange}{$\uparrow$ 16.3\%} ($1.16\times$) \\
    TextMAS (4-Agent) & \res{t_tm_arce} & \res{t_tm_arcc} & \res{t_tm_gsm} & \res{t_tm} & \res{t_out_tm} & \textcolor{relayorange}{$\uparrow$ 257.7\%} ($3.58\times$) \\
    LatentMAS (4-Agent) & \res{t_lmas_arce} & \res{t_lmas_arcc} & \res{t_lmas_gsm} & \res{t_lmas} & \res{t_out_lmas} & Baseline ($1.00\times$) \\
    \rowcolor{ours}[\tabcolsep][\tabcolsep]
    \textbf{\method{} (Ours)} & \textbf{\res{t_ours_arce}} & \textbf{\res{t_ours_arcc}} & \textbf{\res{t_ours_gsm}} & \textbf{\res{t_ours}} & \textbf{\res{t_out_ours}} & \textcolor{relaygreen}{\textbf{$\downarrow$ 20.2\%}} (\textbf{0.80$\times$}) \\
    \bottomrule
  \end{tabular}%
  }
  \caption{\textbf{Generalization to text-based reasoning benchmarks on Qwen3-8B} across 300 test questions each from ARC-Easy, ARC-Challenge, and GSM8K. Operating via Stage~1 early stopping alone, \method{} improves accuracy over 4-agent LatentMAS while reducing decoding tokens by 20.2\%.}
  \label{tab:text}
  \vspace{-3.5mm}
\end{table}

\paragraph{Offline TRPS improves over un-tuned structural heuristics.}
In \cref{tab:main}, the un-tuned Qwen3-VL heuristic triggers visual relay late at 1{,}024 tokens without Quick Check (\texttt{cut1024/DE/S}), allowing unresolved trajectories to accumulate reasoning noise before handover.
TRPS selects an earlier 512-token handover with Quick Check (\texttt{cut512/DE/D1}) on cached development traces, cutting tokens by an additional \textbf{17.9\%} on 8B ($+0.4\%$, CI $[-0.6, +1.4]$) and \textbf{14.1\%} on 4B ($+1.1\%$, CI $[+0.1, +2.0]$).
On InternVL3.5-8B-HF, dev-set threshold calibration at the 512-token handover further cuts tokens by \textbf{10.5\%} ($827 \to 740$) over zero-shot threshold transfer.

\paragraph{Token savings translate into B200 GPU speedups over sequential MAS.}
Measured on NVIDIA B200 GPUs across 91 paired test questions (\cref{fig:exp_summary}c; see appendix for per-benchmark latency), periodic latent readouts add minimal overhead ($23.7$ tokens or $4.5\%$ on 8B; $18.8$ tokens or $4.3\%$ on 4B) because they reuse the Solver's KV cache in a single forward pass.
By cutting long-tail decoding and redundant agent calls, \method{} runs \textbf{1.60$\times$} faster than 4-agent LatentMAS and \textbf{2.59$\times$} faster than 4-agent TextMAS on 8B ($13.31$\,s vs.\ $21.25$\,s and $34.46$\,s), and \textbf{1.37$\times$} faster than LatentMAS and \textbf{2.41$\times$} faster than TextMAS on 4B ($12.61$\,s vs.\ $17.34$\,s and $30.36$\,s), while remaining comparable in latency to a single unstopped Solver (\texttt{CoT+guard}: $14.52$\,s on 8B, $11.84$\,s on 4B; 95\% latency-ratio CIs include $1.0$).

\subsection{Generalization to Text-Based Reasoning Tasks}
\label{sec:text}

\paragraph{Stage 1 early stopping transfers to text-only reasoning.}
On text-only benchmarks where Stage~2 image cropping is inactive (\cref{tab:text}), \method{} operates via Stage~1 early stopping alone.
Just as in multimodal tasks, standard LLMs and sequential MAS continue generating redundant steps after reaching the correct answer.
Halting once latent readouts stabilize improves average accuracy by \textbf{+\res{h3_pt}\%} over 4-agent LatentMAS while cutting decoding tokens by \textbf{20.2\%}, and improves over Single LLM by $+1.6\%$ with $31.4\%$ fewer tokens.
Full 4-agent TextMAS achieves higher text accuracy ($96.2\%$) but requires $4.48\times$ the tokens of \method{} ($2{,}277$ vs.\ $508$).

\subsection{Ablation Study}
\label{sec:ablation}

\begin{table}[t]
  \centering\footnotesize
  \setlength{\tabcolsep}{3.0pt}
  \resizebox{\linewidth}{!}{%
  \begin{tabular}{@{}lccc@{}}
    \toprule
    Method / Ablation Variant (Qwen3-VL-8B) & Acc.\ (\%) & Tokens & Relative Cost \\
    \midrule
    \textbf{\method{} (Full Two-Stage, Ours)} & \textbf{\res{acc_ours}} & \textbf{\res{out_ours}} & \textbf{1.00$\times$} \\
    \midrule
    \multicolumn{4}{@{}l}{\emph{1. Stage 1 \& Stage 2 Component Ablations}} \\
    \quad w/o Stage 1 \& 2 (Solver Only, CoT+guard) & \res{acc_cotg} & \res{out_cotg} & $1.13\times$ \\
    \quad w/o Stage 2 (Stage 1 Early Stop Only) & \res{acc_c1} & \res{out_c1} & $1.07\times$ \\
    \quad w/o Consistency Check (Confidence Only) & 73.4 & 513 & $0.98\times$ \\
    \quad w/o Confidence Filter (Consistency Only) & 69.6 & 336 & $0.64\times$ \\
    \quad Cut at 512 tok w/o Verifier (Naive Cut) & \res{x_cut_acc} & \res{x_cut_out} & $0.63\times$ \\
    \midrule
    \multicolumn{4}{@{}l}{\emph{2. Intervention Timing Comparison}} \\
    \quad Verify Every Run (No Trigger) & \res{acc_always} & \res{out_always} & $1.78\times$ \\
    \quad After-Finish Trigger (Conf.\ Proxy) & \res{x_confm_acc} & \res{x_confm_out} & $1.35\times$ \\
    \quad Early-Stop Trigger (Matched Call Rate) & \res{x_c512_acc} & \res{x_c512_out} & $1.01\times$ \\
    \midrule
    \multicolumn{4}{@{}l}{\emph{3. Visual Relay \& Offline TRPS Components}} \\
    \quad w/o Local Crops (Global Image Verifier) & \res{acc_si} & \res{out_si} & $1.23\times$ \\
    \quad w/o Quick Check (Full Verifier Re-Solve) & \res{fr_acc_dev} & \res{fr_out_dev} & $1.15\times$ \\
    \quad w/o Offline TRPS (Heuristic 1{,}024 tok) & 73.1 & 641 & $1.22\times$ \\
    \bottomrule
  \end{tabular}%
  }
  \caption{\textbf{Stage 1 and Stage 2 component and timing ablations on Qwen3-VL-8B} (921 test questions, 3 seeds). Confidence filtering prevents premature Stage~1 exits, while early Stage~2 handover with Quick Check cuts token cost substantially relative to after-finish triggers and unconditional verification.}
  \label{tab:ablation}
  \vspace{-3mm}
\end{table}

\begin{figure}[t]
  \centering
  \includegraphics[width=0.98\linewidth]{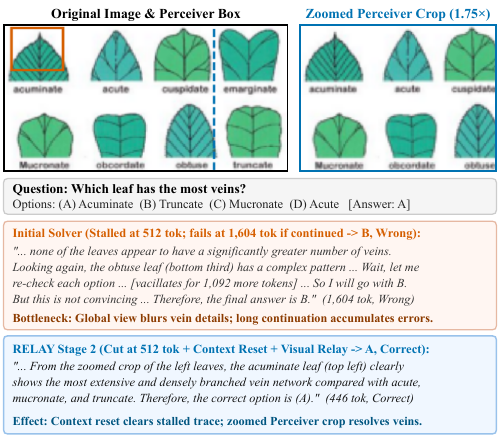}
  \vspace{-2mm}
  \caption{\textbf{Qualitative case of Visual Relay on MathVista (dev \#116).} When the initial Solver stalls on fine leaf-vein details, continuing generation vacillates to a wrong answer (1{,}604 generated tokens); \method{} stops the Solver early, resets the context, and passes a $1.75\times$ Perceiver crop to the Verifier, which resolves the correct answer in 446 generated tokens.}
  \label{fig:case_study}
  \vspace{-3.5mm}
\end{figure}

\begin{table}[t]
  \centering\footnotesize
  \setlength{\tabcolsep}{2.2pt}
  \resizebox{\linewidth}{!}{%
  \begin{tabular}{@{}llcccc@{}}
    \toprule
    Evaluation Setting & Context Reset? & Visual Input & Acc.\ (\%) & Dec. & Proc. \\
    \midrule
    \multirow{2}{*}{\begin{tabular}[c]{@{}l@{}}\textbf{Perception-Error}\\\emph{(12 dev q., 25 states)}\end{tabular}}
    & No (Keep Trace) & Verified Clue (Oracle) & 41.3 & $1.00\times$ & $1.00\times$ \\
    & \textbf{Yes (Reset)} & \textbf{Verified Clue (Oracle)} & \textbf{90.7 (+49.3)} & $1.10\times$ & $1.85\times$ \\
    \midrule
    \multirow{3}{*}{\begin{tabular}[c]{@{}l@{}}\textbf{Standard-Res Relay}\\\emph{(512 tok, 921 test)}\end{tabular}}
    & No (Cut at 512) & None (No Verifier) & \res{x_cut_acc} ($-1.7$) & $0.63\times$ & -- \\
    & Yes (Reset) & Global Image (Indep.) & \res{acc_si} (+0.5) & $1.23\times$ & -- \\
    & \textbf{Yes (Reset)} & \textbf{Perceiver Crops + QC} & \textbf{\res{acc_ours} (Ours)} & \textbf{1.00$\times$} & -- \\
    \midrule
    \multirow{4}{*}{\begin{tabular}[c]{@{}l@{}}\textbf{High-Res Dev Sets}\\\emph{(V*Bench, HR-Bench,}\\\emph{MME-RW-Lite)}\end{tabular}}
    & No (Keep Trace) & Global (2-Ag.\ LatentMAS) & 69.9 & $1.00\times$ & $1.00\times$ \\
    & No (Keep Trace) & Perceiver Local Crops & 69.9 (+0.0) & $1.78\times$ & $0.43\times$ \\
    & Yes (Reset) & Random Local Crops & 65.9 ($-3.9$) & $1.07\times$ & $0.38\times$ \\
    & \textbf{Yes (Reset)} & \textbf{Perceiver Crops} & \textbf{74.0 (+4.1)} & $1.04\times$ & \textbf{0.39$\times$} \\
    \bottomrule
  \end{tabular}%
  }
  \caption{\textbf{Factor analysis of Context Reset and Perceiver Local Crops} across the oracle perception-error diagnostic, standard-resolution test relay, and high-resolution development sets (40 V*Bench, 40 HR-Bench averaged across 4K/8K and 4 option rotations, 300 MME-RealWorld-Lite; see appendix for per-benchmark results). Dec.\ and Proc.\ denote relative decoding and shared-root processed token ratios within each group.}
  \label{tab:hires_clean}
  \vspace{-3.5mm}
\end{table}

\paragraph{Confidence filtering is critical for reliable Stage 1 early stopping.}
In \cref{tab:ablation} (Group~1; see appendix), Stage~1 alone cuts token cost relative to Solver Only ($564$ vs.\ $596$ tokens, $-0.4\%$ accuracy, CI $[-0.8, -0.1]$).
Removing the confidence filter (consistency only) drops accuracy sharply by \textbf{3.9\%} ($73.5\% \to 69.6\%$), because an uncertain Solver can repeat the same tentative guess across early checkpoints before finishing its reasoning.
Confidence filtering eliminates these premature exits and yields a broad plateau across checkpoint pairs (\cref{fig:exp_summary}b).
Removing consistency checking alone leaves test accuracy nearly unchanged ($73.4\%$, $-0.04\%$, CI $[-0.19, +0.11]$) at $0.98\times$ cost; the frozen policy retains two-checkpoint agreement alongside confidence filtering.

\paragraph{Early handover balances accuracy and token cost against naive truncation and after-finish triggers.}
In \cref{tab:ablation} (Groups~1--2) and \cref{fig:exp_summary}a, simply truncating unresolved runs at 512 tokens without verification (Naive Cut) causes a \textbf{\res{x_cut_loss_pt}\%} accuracy drop ($71.8\%$), showing that unconverged trajectories benefit from visual re-evaluation.
Conversely, waiting until the Solver completes its full trajectory before triggering verification (completion-confidence After-Finish Trigger proxy: $74.2\%$, $710$ tokens) or verifying every input unconditionally ($74.6\%$, $937$ tokens) raises token cost to $1.35\times$--$1.78\times$.
\method{} uses \textbf{25.9\% fewer tokens} than the after-finish proxy ($526$ vs.\ $710$ tokens) with comparable accuracy ($73.5\%$ vs.\ $74.2\%$).

\begin{figure*}[t]
  \centering
  \includegraphics[width=0.96\textwidth]{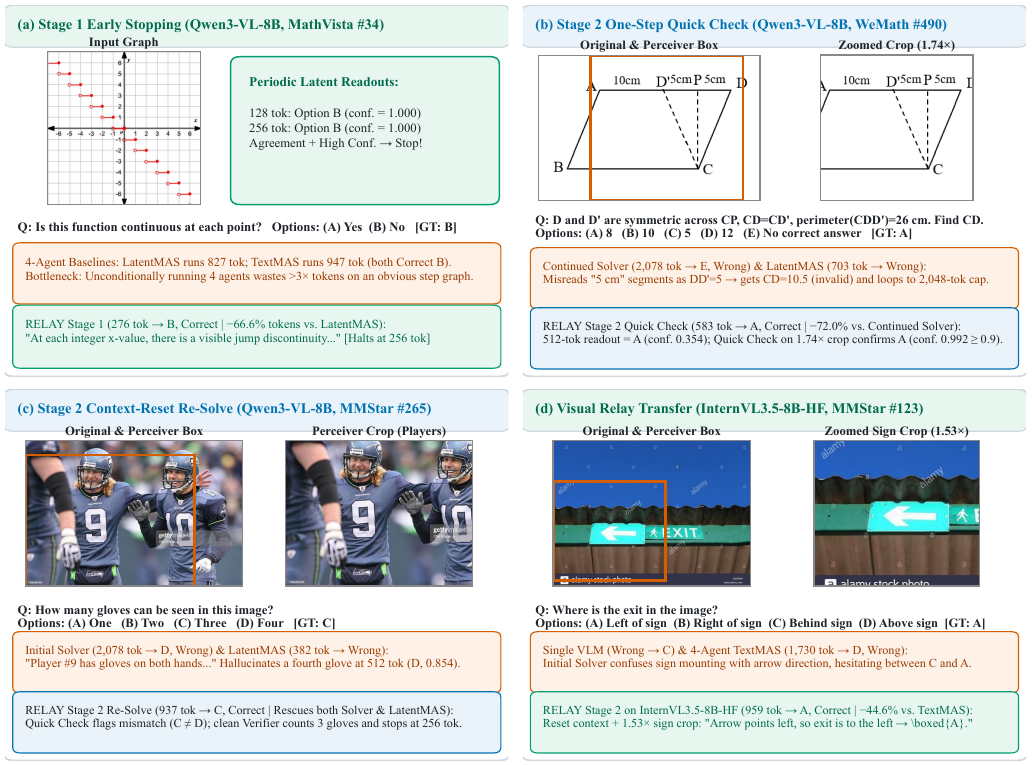}
  \vspace{-3.5mm}
  \caption{\textbf{Verified test-set qualitative examples across \method{} execution paths and VLM backbones.}
    \emph{(a) Stage 1 Early Stopping (MathVista \#34, Qwen3-VL-8B):} Halts at 256 tokens ($276$ total) once consecutive readouts agree at $1.000$ confidence, saving $66.6\%$ of tokens over 4-agent LatentMAS ($827$ tokens).
    \emph{(b) Stage 2 Quick Check (WeMath \#490, Qwen3-VL-8B):} Zooms $1.74\times$ into $\triangle CDD'$ to confirm the Solver's tentative answer ($0.354 \to 0.992$) in one forward pass ($583$ tokens), preventing a 2{,}078-token cap failure.
    \emph{(c) Stage 2 Context-Reset Re-Solve (MMStar \#265, Qwen3-VL-8B):} Discards the Solver's hallucinated fourth-glove trace and counts three visible gloves cleanly ($937$ tokens), rescuing continued Solver and LatentMAS errors.
    \emph{(d) Visual Relay Transfer (MMStar \#123, InternVL3.5-8B-HF):} Crops $1.53\times$ onto the exit sign so the clean Verifier reads the left arrow ($959$ tokens), correcting Single VLM and 4-agent TextMAS ($1{,}730$ tokens).}
  \label{fig:qual-cases}
  \vspace{-4.5mm}
\end{figure*}

\paragraph{Quick Check and Offline TRPS reduce verification overhead.}
In \cref{tab:ablation} (Group~3; see appendix for full structural and threshold tables), Quick Check reduces decoding tokens by \textbf{12.7\%} ($526$ vs.\ $603$ tokens) with similar observed accuracy ($73.5\%$ vs.\ $73.5\%$, CI $[-0.58, +0.64]$), while replacing local crops with an independent-seed global-image Verifier increases token cost by $23\%$ ($647$ tokens, $74.0\%$ accuracy).
Because context reset decouples verification from the Solver's stopping step, Offline TRPS replays all 16 cached threshold configurations across 7 cross-dataset and cross-scale development tasks in \textbf{1.38 seconds} (selector stage, without extra VLM rollouts).
In exploratory replay on the cached development sets, cross-dataset and cross-scale threshold selection reduces decoder cost by \textbf{25\%--31\%} while accuracy intervals remain wide ($[-1.8, +1.7]$ on 4B; $[-2.2, +1.9]$ on 8B; see appendix).

\subsection{Visual Relay Mechanism and Case Study}
\label{sec:stops}

\begin{table}[t]
  \centering\footnotesize
  \setlength{\tabcolsep}{2.4pt}
  \resizebox{\linewidth}{!}{%
  \begin{tabular}{@{}lccccc@{}}
    \toprule
    & & \multicolumn{2}{c}{\textbf{\method{} (Ours)}} & \multicolumn{2}{c}{\textbf{Solver Continues}} \\
    \cmidrule(lr){3-4}\cmidrule(lr){5-6}
    Execution Path & Share & Acc.\ (\%) & Tokens & Acc.\ (\%) & Tokens \\
    \midrule
    1. Stage 1 Early Stop (256 tok) & \res{path_share_settled}\% & \res{path_acc_settled} & 276 & -- & -- \\
    2. Solver Completion w/o Handover & \res{path_share_finish}\% & \res{path_acc_finish} & 261 & -- & -- \\
    \midrule
    3. Stage 2 Quick-Check Exit & \res{w_glance_share}\% & \textbf{\res{w_glance_ours_acc}} & \textbf{\res{w_glance_ours_out}} & \res{w_glance_c1_acc} & \res{w_glance_c1_out} \\
    4. Stage 2 Clean Re-Solve & \res{fresh_solve_share}\% & 55.3 & 1{,}709 & 56.4 & 1{,}515 \\
    \midrule
    All Relayed Unresolved Runs (512 tok) & \res{w_all_share}\% & \res{w_all_ours_acc} & \textbf{\res{w_all_ours_out}} & \res{w_all_c1_acc} & \res{w_all_c1_out} \\
    \bottomrule
  \end{tabular}%
  }
  \vspace{-2mm}
  \caption{\textbf{Execution path breakdown of \method{} on Qwen3-VL-8B} (2{,}763 test runs; omits $0.4\%$ high-resolution overview exits). Over three quarters of runs exit via the Solver alone, while Stage~2 matches continued-Solver accuracy on relayed runs at 10.5\% lower token cost.}
  \label{tab:stops}
  \vspace{-3.5mm}
\end{table}

\paragraph{Context reset and localized Perceiver crops are both essential on high-resolution tasks.}
\Cref{tab:hires_clean} disentangles context reset from localized visual crops.
On the 12-question oracle diagnostic, supplying the same verified visual clue in a reset context raises accuracy from $41.3\%$ to \textbf{90.7\%} ($+49.3\%$, at $1.10\times$ decoding tokens).
On the high-resolution development sets, providing Perceiver crops while keeping the Solver's trace brings no gain over 2-agent global-image LatentMAS ($69.9\%$ vs.\ $69.9\%$, at $1.78\times$ decoding cost), whereas resetting the context with random crops degrades accuracy to $65.9\%$ ($-3.9\%$).
Combining context reset with Perceiver crops reaches \textbf{74.0\%} macro accuracy ($+4.1\%$ over 2-agent LatentMAS and $+8.0\%$ over random crops) while cutting shared-root processed visual and text tokens to \textbf{0.39$\times$} (at $1.04\times$ decoding tokens), with especially strong gains on V*Bench (\textbf{89.2\%}; see appendix).

\paragraph{Execution path breakdown shows how the two stages complement each other.}
As shown in \cref{tab:stops}, \textbf{75.3\%} of test runs exit via the Solver alone, either through Stage~1 early stopping ($28.0\%$, $78.8\%$ accuracy, $276$ tokens) or natural completion without handover ($47.3\%$, $77.9\%$ accuracy, $261$ tokens).
For the remaining \textbf{\res{w_all_share}\%} of unconverged runs relayed at 512 tokens, Stage~2 matches continued-Solver accuracy ($58.6\%$ vs.\ $58.7\%$) while reducing average decoding cost by \textbf{10.5\%} ($1{,}342$ vs.\ $1{,}499$ tokens).
Within Stage~2, one-step Quick Check resolves nearly one third of relayed runs at \textbf{40\%} of the continuation token cost ($589$ vs.\ $1{,}466$ tokens) while gaining $+1.8\%$ accuracy ($65.3\%$ vs.\ $63.5\%$), reserving full Verifier re-solving for the remaining hard cases (\cref{fig:case_study,fig:qual-cases}).

\section{Conclusion}
\label{sec:conclusion}

We presented \textbf{\method{}} (\emph{\methodlong{}}), a training-free early-stopping framework for multimodal multi-agent reasoning.
We observe that VLMs are prone to overthinking on easy visual questions, which wastes tokens and can degrade accuracy, while on hard questions early perceptual errors accumulate and trap agents in unproductive reasoning loops.
To address both issues, \method{} combines Readout-Driven Early Stopping to halt converged runs early with Context-Reset Visual Relay to re-evaluate unconverged runs on localized image crops in a clean context, and uses \emph{Offline Trace-Replay Policy Search (TRPS)} to select stopping thresholds offline without extra VLM inference.
Experiments across multimodal, high-resolution visual, and text benchmarks show that \method{} improves accuracy over fixed sequential MAS while substantially reducing decoding token cost and inference latency across model scales.

\paragraph{Limitations and future work.}
Despite these gains, \method{} has three main limitations.
First, Stage~1 relies directly on the frozen VLM's raw softmax confidence, which in rare cases can stop prematurely on overconfident arithmetic shortcuts before a later self-correction.
Second, Stage~2 uses a uniform context margin when cropping regions, which works well on natural images but can occasionally clip outer coordinate axes on scientific plots.
Third, zero-shot bounding-box localization is sensitive to model-specific JSON formatting conventions, causing high fallback rates to global-image re-solving on transferred visual backbones such as InternVL3.5-8B-HF.
Integrating lightweight confidence calibration, diagram-aware crop margins, and model-adaptive localization prompts are promising directions for future work.

{
  \small
  \bibliographystyle{ieeenat_fullname}
  \bibliography{refs}
}

\maketitleappendix
\appendix
\section{Algorithmic Formulation and Implementation Details}
\label{app:impl}

\begin{algorithm}[t]
  \caption{\textbf{\method{}: Early Stopping, Visual Relay, and Offline TRPS.}}
  \label{alg:relay_short}
  \footnotesize
  \begin{algorithmic}[1]
    \Statex \textbf{Offline TRPS (Once on Cached Dev Outputs):}
    \State Replay cached Solver readouts and clean-context Verifier outputs across candidate stopping steps and confidence thresholds; lock policy $\theta^\star = (t_1, t_2, h, \tau_s, \tau_g)$ for test inference (\cref{eq:tradeoff_formulation}).
    \vspace{2pt}
    \Statex \textbf{Online Inference (Given Image $I$ and Question $q$):}
    \State Encode image once into Shared Visual Cache $R$; initialize Solver context $H_S = R \oplus q$.
    \Statex \emph{// Stage 1: Readout-Driven Early Stopping}
    \While{Solver generates tokens (up to stopping step $h$)}
      \If{Solver naturally finishes with an answer} \Return answer \EndIf
      \State At steps $t \in \{t_1, t_2, h\}$, apply $\operatorname{Readout}(C_t)$ on a temporary KV branch to get tentative answer $\hat{a}_t$ and confidence $c_t$.
      \If{consecutive readouts agree ($\hat{a}_{t_1} = \hat{a}_{t_2}$) and confidence $c_{t_2} \ge \tau_s$}
        \State \textbf{Stop early} and \Return $\hat{a}_{t_2}$.
      \EndIf
    \EndWhile
    \Statex \emph{// Stage 2: Context-Reset Visual Relay (Triggered at Step $h$)}
    \State Record tentative answer $\hat{a}_h$, \textbf{stop Solver}, and \textbf{reset context} (discard Solver text).
    \State Crop task-relevant regions $C_{\mathrm{crop}}$ via Perceiver and form clean Verifier context $H_V = R \oplus C_{\mathrm{crop}} \oplus q$.
    \State Apply one-step $\operatorname{Readout}$ (\emph{Quick Check}) on $H_V$ to obtain $(\hat{a}_V, c_V)$:
    \If{$\hat{a}_V = \hat{a}_h$ and $c_V \ge \tau_g$} \Return $\hat{a}_h$ \EndIf
    \State \Return fresh answer generated by Verifier from scratch on $H_V$.
  \end{algorithmic}
\end{algorithm}

\paragraph{Two-step Offline TRPS formulation and selection rule.}
\Cref{alg:relay_short} summarizes the online inference and offline threshold selection procedure of \method{}.
Let $\theta = (t_1, t_2, h, \tau_s, \tau_g) \in \Theta$ denote the Stage~1 readout steps, Stage~2 handover step, and confidence thresholds.
Given cached development traces $\mathcal{D}_{\mathrm{dev}}$, Offline Trace-Replay Policy Search (TRPS, \cref{sec:solver}) maximizes development accuracy subject to a 95\% bootstrap upper confidence bound $\widehat{U}_{0.95}$ on the relative token cost ratio $\rho_{\mathrm{tok}}$ (breaking ties by lower cost):
\begin{equation}
  \begin{aligned}
    \theta^\star = \arg\max_{\theta \in \Theta}\; & \operatorname{Acc}(\theta; \mathcal{D}_{\mathrm{dev}}) \\
    \text{s.t.}\; & \widehat{U}_{0.95}\!\big(\rho_{\mathrm{tok}}(\theta)\big) \le \gamma_{\mathrm{tok}}.
  \end{aligned}
  \label{eq:tradeoff_formulation}
\end{equation}
Under $\gamma_{\mathrm{tok}} = 0.95$ vs.\ 4-agent LatentMAS on Qwen3-VL-8B, Step~1 evaluates structural horizons $h \in \{0, 512, 1024, 2048\}$ and verification inputs (including Stage~1 only); note that late relay at $h=1024$ (\texttt{cut1024/DE/D1}) has point cost ratio $0.90\times$ but 95\% upper bound $0.986 > 0.95$, so it is infeasible under \cref{eq:tradeoff_formulation}, and Step~1 selects $h=512$ with Perceiver crops and Quick Check (\texttt{cut512/DE/D1}).
Step~2 sweeps $(\tau_s, \tau_g)$ (\cref{app:offline-selector}), locking $\theta^\star = (128, 256, 512, 0.9982, 0.9)$ on Qwen3-VL-8B/4B ($\tau_s{=}0.99, \tau_g{=}1.0$ frozen on InternVL3.5-8B-HF dev traces prior to test inference).

\paragraph{Latent readout probing on multiple-choice and free-form tasks.}
Each probe $\operatorname{Readout}(C_t)$ (\cref{eq:readout}) branches a temporary KV copy of $C_t$ and discards it right after, leaving ongoing generation unchanged.
On multiple-choice tasks (MMStar, WeMath, MathVista, ARC), we append ``\texttt{Therefore, the final answer is \textbackslash boxed\{}'' (10 tokens on Qwen3-VL; 11 on InternVL3.5-8B-HF) in Stage~1 or ``\texttt{The final answer is \textbackslash boxed\{}'' (7 tokens on Qwen3-VL; 8 on InternVL3.5-8B-HF) in Quick Check, taking $\hat{a}_t = \arg\max_{o \in \mathcal{O}} p_t(o)$ and $c_t = p_t(\hat{a}_t)$ in one forward pass.
On open-ended items (GSM8K and free-form tasks), the probe greedily decodes up to 24 tokens until ``\texttt{\}}'' and uses the product of answer-token probabilities as $c_t$.
Two readouts agree when their normalized boxed answers match.

\paragraph{Ultra-high-resolution inputs and fair token accounting.}
Standard images ($\le 2{,}048$ visual tokens) are encoded once into $R$ (\cref{alg:relay_short}); for ultra-high-resolution inputs ($>2{,}048$ visual tokens, \res{path_share_rstop}\% of runs), $R$ encodes a downsampled overview, the Perceiver crops $C_{\mathrm{crop}}$, and a confident Quick Check ($\ge 0.9$) exits early.
Across methods, $C_{\mathrm{tok}}$ counts active generated text tokens, readout prompt tokens under each model's native tokenizer, and 60 multimodal latent steps for LatentMAS (120 on text); even excluding latent steps, \method{} cuts 8B LatentMAS token cost to $0.84\times$ (\cref{tab:config}).

\begin{table}[t]
  \centering\footnotesize
  \setlength{\tabcolsep}{2.5pt}
  \resizebox{\linewidth}{!}{%
  \begin{tabular}{@{}lllll@{}}
    \toprule
    Method & Agents & Role Prompts & Latent Steps & Max Tokens / Agent \\
    \midrule
    \textbf{\method{} (Solver, Verifier)} & 1--3 & Zero-shot CoT & -- & 2{,}048 \\
    \textbf{\method{} (Perceiver)} & -- & Box Locator & -- & 128 \\
    Solver Only; Self-Consistency (3 Chains) & 1; 3 & Zero-shot CoT & -- & 2{,}048 \\
    Single VLM (Default) & 1 & Standard VLM & -- & 4{,}096 \\
    LatentMAS (4-Agent, Default) & 4 & Sequential MAS & 20 / role & 4{,}096 \\
    LatentMAS (2-Agent; 4-Role Prompts) & 2; 4 & Role-specific & 20 / role & 4{,}096 \\
    TextMAS (4-Agent) & 4 & Sequential Text & -- & 4{,}096 \\
    \bottomrule
  \end{tabular}%
  }
  \caption{\textbf{Hyperparameter settings for \method{} and compared baselines.}
    All multimodal methods use identical sampling parameters (temperature $0.6$, top-$p=0.95$, top-$k=20$) across Qwen3-VL-8B, Qwen3-VL-4B, and InternVL3.5-8B-HF. Text benchmarks use a 2{,}048-token cap across all methods.}
  \label{tab:config}
\end{table}

\paragraph{Zero-shot role prompt templates.}
All three agent roles share one frozen VLM with images placed before text:
\begin{itemize}[leftmargin=*,itemsep=1.5pt,topsep=1.5pt]
  \item \textbf{Solver:} ``\texttt{You are a helpful assistant. Question: \{question\}\textbackslash n\textbackslash nAnswer the question from the original image. Check the decisive visual evidence and the strongest competing answer. You must reason step-by-step without outputting other irrelevant information and then output the final answer inside \textbackslash boxed\{YOUR\_FINAL\_ANSWER\}.}''
  \item \textbf{Perceiver:} ``\texttt{Locate the part of the image that must be looked at to answer this question: \{question\_stem\}\textbackslash nIf two separate objects must be compared, output one box for each. Output the bounding boxes in JSON format: \{"bbox\_2d": [x1, y1, x2, y2]\}, with coordinates relative to the image on a 0-1000 scale.}'' (up to two boxes if parsed union area $\le 50\%$, expanded $3\times$ and resized toward $[448, 1280]$\,px subject to a $2\times$ upscaling cap and visual token budget).
  \item \textbf{Verifier:} Prepends ``\texttt{The second image is an enlarged view of the region of the first image that is relevant to the question.}'' to the Solver prompt in a reset context.
\end{itemize}

\paragraph{Benchmark splits and evaluation protocol.}
The multimodal development set contains 367 standard-resolution items on 8B (117 MMStar, 100 MathVista, 150 WeMath, formed by excluding two high-resolution MMStar items with $>2{,}048$ visual tokens from the 369-item pilot pool used on 4B).
The frozen test set comprises 921 non-overlapping questions across MMStar (354), MathVista (201), and WeMath (366), after filtering 34 WeMath items sharing question groups with the development split; 453 items were never run prior to confirmation (Tier~A: $+4.3\%$ vs.\ LatentMAS) and 468 items had one unanalyzed earlier V2 output (Tier~B: $+2.0\%$, 95\% CI $[-0.1, +4.2]$).
Qwen3-VL is evaluated over 3 seeds and InternVL3.5-8B-HF over 1 paired seed; text evaluation uses official test indices 300--599 (300 items each) from ARC-Easy, ARC-Challenge, and GSM8K over 3 seeds.

\section{Additional Experimental Results}
\label{app:more}

\subsection{Adaptive Routing, Multi-Seed Stability, and Baseline Upgrades}
\label{app:tiers}

\begin{figure*}[t]
  \centering
  \includegraphics[width=\textwidth]{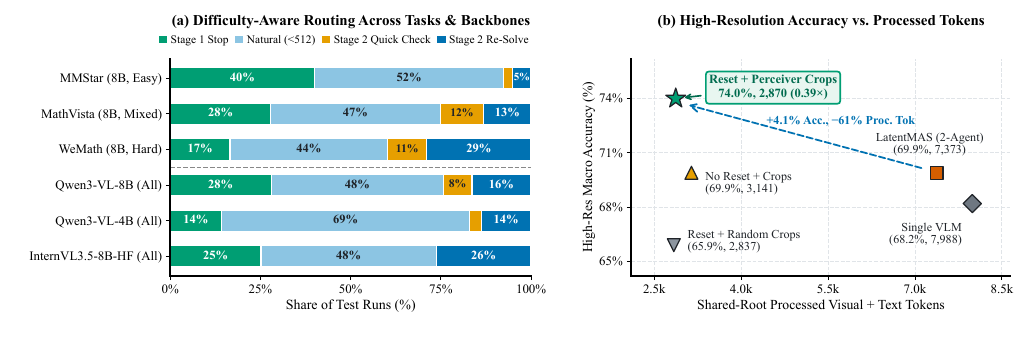}
  \vspace{-3mm}
  \caption{\textbf{Difficulty-aware execution routing and high-resolution accuracy--cost trade-off.}
    \emph{(a) Execution path shares:} Under a single locked policy, \method{} exits via the Solver alone on $92.4\%$ of general perception items (MMStar) while escalating $39.7\%$ of multi-step geometry items (WeMath) to Stage~2 visual relay, and adapts path usage across Qwen3-VL-8B, Qwen3-VL-4B, and InternVL3.5-8B-HF.
    \emph{(b) High-resolution accuracy vs.\ shared-root processed tokens:} Across V*Bench, HR-Bench, and MME-RealWorld-Lite development sets, Reset + Perceiver Crops achieves the highest macro accuracy ($74.0\%$) while cutting shared-root processed tokens by $61\%$ ($0.39\times$ vs.\ 2-agent LatentMAS).}
  \label{fig:supp-routing-highres}
\end{figure*}

\begin{table}[t]
  \centering\footnotesize
  \setlength{\tabcolsep}{2.4pt}
  \resizebox{\linewidth}{!}{%
  \begin{tabular}{@{}llccccc@{}}
    \toprule
    Group & Benchmark / Backbone & Stage 1 Stop & Natural ($<$512) & Quick Check & Clean Re-Solve & Stage 2 Total \\
    \midrule
    \multirow{3}{*}{\textbf{By Task (8B)}}
    & MMStar ($N{=}354$) & 40.0\% & 52.4\% & 2.5\% & 5.1\% & \textbf{7.6\%} \\
    & MathVista ($N{=}201$) & 27.9\% & 46.9\% & 12.1\% & 13.1\% & \textbf{25.2\%} \\
    & WeMath ($N{=}366$) & 16.6\% & 43.7\% & 10.8\% & 28.9\% & \textbf{39.7\%} \\
    \midrule
    \multirow{3}{*}{\textbf{By Backbone}}
    & Qwen3-VL-8B (All) & 28.0\% & 47.7\% & 7.9\% & 16.3\% & 24.2\% \\
    & Qwen3-VL-4B (All) & 14.3\% & 68.6\% & 3.3\% & 13.8\% & 17.1\% \\
    & InternVL3.5-8B-HF (All) & 25.2\% & 48.5\% & 0.0\% & 26.3\% & 26.3\% \\
    \bottomrule
  \end{tabular}%
  }
  \caption{\textbf{Execution path distribution across benchmarks and VLM backbones} on the 921-question test set (Natural includes $0.4\%$ overview exits on Qwen3-VL and $0.8\%$ cap exits on InternVL3.5-8B-HF). \method{} automatically scales Stage~2 visual relay from $7.6\%$ on MMStar to $39.7\%$ on WeMath.}
  \label{tab:routing_breakdown}
\end{table}

\paragraph{\method{} scales Stage 2 visual relay with task difficulty.}
Extending \cref{tab:stops}, \cref{fig:supp-routing-highres}a and \cref{tab:routing_breakdown} break down execution paths across tasks under one frozen policy on Qwen3-VL-8B.
On general perception in MMStar, $92.4\%$ of runs exit via the Solver alone ($40.0\%$ Stage~1 stop at 276 tokens; $52.4\%$ natural finish at 234 tokens), invoking Stage~2 on only $7.6\%$ of runs.
On multi-step geometry in WeMath, Stage~2 invocation rises over fivefold to $39.7\%$ ($10.8\%$ Quick Check; $28.9\%$ Clean Re-Solve), confirming automatic difficulty-aware allocation.

\paragraph{Execution routing adapts across VLM backbones.}
Across model scales and families (\cref{tab:routing_breakdown}, bottom), Qwen3-VL-4B finishes shorter Solver chains naturally more often ($68.6\%$) while relaying $17.1\%$ of runs to Stage~2 ($3.3\%$ Quick Check, $13.8\%$ Clean Re-Solve).
On InternVL3.5-8B-HF ($\tau_g{=}1.0$), $25.2\%$ of queries exit at Stage~1 and all $26.3\%$ of unconverged runs proceed directly to clean Verifier re-solving ($0.0\%$ Quick Check).

\begin{table}[t]
  \centering\footnotesize
  \setlength{\tabcolsep}{2.4pt}
  \resizebox{\linewidth}{!}{%
  \begin{tabular}{@{}llccccc@{}}
    \toprule
    Backbone & Method & Seed 1 & Seed 2 & Seed 3 & Mean Acc. & Avg.\ Tokens \\
    \midrule
    \multirow{4}{*}{\textbf{Qwen3-VL-8B}}
    & Single VLM (Default) & 73.2 & 71.9 & 72.2 & 72.4 & 529 \\
    & TextMAS (4-Agent) & 68.2 & 68.4 & 68.0 & 68.2 & 1{,}373 \\
    & LatentMAS (4-Agent) & 69.5 & 70.4 & 71.1 & 70.3 & 689 \\
    \rowcolor{ours}
    \cellcolor{white} & \textbf{\method{} (Ours)} & \textbf{71.7} & \textbf{73.1} & \textbf{75.7} & \textbf{73.5} & \textbf{526} \\
    \midrule
    \multirow{4}{*}{\textbf{Qwen3-VL-4B}}
    & Single VLM (Default) & 70.8 & 70.9 & 70.8 & 70.8 & 580 \\
    & TextMAS (4-Agent) & 66.9 & 67.9 & 68.3 & 67.7 & 1{,}292 \\
    & LatentMAS (4-Agent) & 69.8 & 68.7 & 69.3 & 69.3 & 723 \\
    \rowcolor{ours}
    \cellcolor{white} & \textbf{\method{} (Ours)} & \textbf{73.6} & \textbf{72.0} & \textbf{72.4} & \textbf{72.7} & \textbf{441} \\
    \bottomrule
  \end{tabular}%
  }
  \caption{\textbf{Multi-seed stability across all three sampling seeds} on the 921-question test set (temperature $0.6$). \method{} outperforms 4-agent LatentMAS and TextMAS on every individual seed across both Qwen3-VL-8B and Qwen3-VL-4B.}
  \label{tab:seed_stability}
\end{table}

\begin{table}[t]
  \centering\footnotesize
  \setlength{\tabcolsep}{2.4pt}
  \resizebox{\linewidth}{!}{%
  \begin{tabular}{@{}llcccc@{}}
    \toprule
    Backbone & Baseline / Control & Acc.\ (\%) & $C_{\mathrm{tok}}$ & \method{} $\Delta$Acc & \method{} Cost Ratio \\
    \midrule
    8B & LatentMAS (4-Agent, 20 latent steps) & \res{acc_lmas} & \res{out_lmas} & \textbf{+\res{h1_pt}\%} & \textbf{0.76$\times$} \\
    8B & LatentMAS (4-Agent, 40 latent steps) & \res{st_acc_m40} & \res{st_dec_m40} & +3.0\% & 0.69$\times$ \\
    8B & LatentMAS (4-Agent, 80 latent steps) & \res{st_acc_m80} & \res{st_dec_m80} & +2.9\% & 0.62$\times$ \\
    8B & LatentMAS (4-Agent) + Cap Readout & \res{gr_acc_lmas} & \res{gr_dec_lmas} & +1.0\% & 0.92$\times$ \\
    8B & LatentMAS (2-Agent) + Cap Readout & \res{gr_acc_pj} & \res{gr_dec_pj} & +0.4\% & 0.91$\times$ \\
    8B & LatentMAS (4-Role) + Cap Readout & \res{gr_acc_pcrj} & \res{gr_dec_pcrj} & +0.4\% & 0.90$\times$ \\
    \midrule
    4B & LatentMAS (4-Agent, 20 latent steps) & \res{m2_acc_lmas} & \res{m2_out_lmas} & \textbf{+\res{m2_h5_pt}\%} & \textbf{0.61$\times$} \\
    4B & LatentMAS (4-Agent) + Cap Readout & \res{m2_gr_acc_lmas} & \res{m2_gr_dec_lmas} & +0.9\% & 0.79$\times$ \\
    4B & LatentMAS (2-Agent) + Cap Readout & \res{m2_gr_acc_pj} & \res{m2_gr_dec_pj} & +0.9\% & 0.84$\times$ \\
    4B & LatentMAS (4-Role) + Cap Readout & \res{m2_gr_acc_pcrj} & \res{m2_gr_dec_pcrj} & +1.6\% & 0.80$\times$ \\
    \midrule
    InternVL3.5 & Single VLM (Default) & 70.5 & 194 & +2.0\% & 3.81$\times$ \\
    InternVL3.5 & Single VLM + Cap Readout & 70.5 & 194 & +2.0\% & 3.81$\times$ \\
    InternVL3.5 & TextMAS (4-Agent, Default) & 69.3 & 3{,}376 & \textbf{+3.1\%} & \textbf{0.22$\times$} \\
    InternVL3.5 & TextMAS (4-Agent) + Cap Readout & 73.1 & 3{,}377 & $-0.7\%$ & \textbf{0.22$\times$} \\
    InternVL3.5 & \method{} (Zero-Shot Transfer) & 71.9 & 827 & +0.5\% & 0.89$\times$ \\
    \rowcolor{ours}
    InternVL3.5 & \textbf{\method{} (Dev-Calibrated)} & \textbf{72.4} & \textbf{740} & -- & 1.00$\times$ \\
    \bottomrule
  \end{tabular}%
  }
  \caption{\textbf{Robustness across LatentMAS latent depths, cap answer-readout controls, and full 6-row InternVL3.5-8B-HF transfer} (921 test items; Qwen3-VL 3 seeds, InternVL3.5-8B-HF 1 paired seed). Scaling latent depth inflates cost with accuracy between 70.3\% and 70.6\%. Against cap-readout controls, \method{} retains consistent decoding-token reductions ($0.22\times$--$0.92\times$) while accuracy differences have 95\% CIs spanning zero.}
  \label{tab:hyp}
\end{table}

\paragraph{\method{} outperforms sequential MAS across every random seed.}
\Cref{tab:seed_stability} evaluates seed-level stability across all three sampling seeds (temperature $0.6$) on the 921-question test set.
\method{} beats 4-agent LatentMAS and TextMAS on every seed on both Qwen3-VL-8B ($+2.2\%$ to $+4.6\%$ over LatentMAS) and Qwen3-VL-4B ($+3.1\%$ to $+3.8\%$), confirming stability under stochastic decoding.

\paragraph{Latent depth, cap-readout controls, and InternVL3.5-8B-HF transfer boundaries.}
In \cref{tab:hyp}, scaling LatentMAS to 40 or 80 latent steps per role inflates cost to 767--854 tokens with accuracy between $70.3\%$ and $70.6\%$ (95\% CIs vs.\ 20 steps include zero).
Appending a cap answer readout (+Cap Readout) recovers invalid cap runs, narrowing accuracy differences so that 95\% CIs vs.\ \method{} span zero on Qwen3-VL (8B: $+1.0\%$ $[-0.8, +2.8]$; 4B: $+0.9\%$ $[-0.6, +2.5]$), while \method{} consistently cuts decoding tokens ($0.79\times$--$0.92\times$).

On InternVL3.5-8B-HF (which retains a Qwen3 text decoder, evaluated over 1 paired seed), appending cap readout to all 69 cap-truncated TextMAS Judge runs eliminates all 67 invalid outputs and raises TextMAS accuracy from $69.3\%$ to $73.1\%$; dev-calibrated \method{} ($72.4\%$, $740$ tokens) achieves comparable accuracy ($-0.65\%$, 95\% CI $[-3.23, +2.00]$) while cutting token cost by $78.1\%$ ($0.22\times$, CI $[0.20, 0.24]$).
Relative to Single VLM (+Cap Readout: $70.5\%$, $194$ tokens), \method{} uses $3.81\times$ tokens (CI $[3.36, 4.30]$) with $+1.95\%$ accuracy (CI $[-0.76, +4.62]$).
Auditing raw InternVL3.5-8B-HF traces reveals that without model-specific prompt adaptation, the Perceiver fails to emit a valid JSON box on $634/921$ items ($68.8\%$; $597/634$ write verbose explanations hitting the 128-token cap), causing $198/242$ ($81.8\%$) of Stage~2 handovers to fall back to global-image clean re-solving.

\subsection{High-Resolution Per-Benchmark and Resolution Breakdown}

\begin{table*}[t]
  \centering\footnotesize
  \setlength{\tabcolsep}{3.8pt}
  \resizebox{0.96\textwidth}{!}{%
  \begin{tabular}{@{}lcccccccccc@{}}
    \toprule
    & \multicolumn{2}{c}{\textbf{V*Bench} ($N{=}40$)} & \multicolumn{2}{c}{\textbf{HR-Bench 4K} ($N{=}40$)} & \multicolumn{2}{c}{\textbf{HR-Bench 8K} ($N{=}40$)} & \multicolumn{2}{c}{\textbf{MME-RW-Lite} ($N{=}300$)} & \multicolumn{2}{c}{\textbf{3-Benchmark Macro}} \\
    \cmidrule(lr){2-3}\cmidrule(lr){4-5}\cmidrule(lr){6-7}\cmidrule(lr){8-9}\cmidrule(lr){10-11}
    Method (Qwen3-VL-8B) & Acc. & Proc.\ Tok & Acc. & Proc.\ Tok & Acc. & Proc.\ Tok & Acc. & Proc.\ Tok & Acc. & Proc.\ Tok ($C_{\mathrm{tok}}$) \\
    \midrule
    Single VLM (Default View) & 77.5 & 3{,}670 & 75.0 & 14{,}218 & 70.9 & 16{,}706 & 54.0 & 4{,}833 & 68.2 & 7{,}988 (274) \\
    LatentMAS (2-Agent, P--J, 4096) & 78.3 & 7{,}127 & \textbf{78.8} & 8{,}866 & \textbf{76.6} & 8{,}769 & 53.7 & 6{,}175 & 69.9 & 7{,}373 (336) \\
    No Reset + Perceiver Crops & 82.5 & 2{,}835 & 77.5 & 3{,}519 & 70.3 & 3{,}728 & 53.3 & 2{,}962 & 69.9 & 3{,}141 (599) \\
    Reset + Random Crops & 76.7 & 2{,}670 & 75.9 & 3{,}123 & 68.8 & 3{,}200 & 48.8 & 2{,}680 & 65.9 & 2{,}837 (361) \\
    \rowcolor{ours}\textbf{Reset + Perceiver Crops (Comp.)} & \textbf{89.2} & \textbf{2{,}686} & 77.8 & \textbf{3{,}155} & 76.3 & \textbf{3{,}308} & \textbf{55.7} & \textbf{2{,}693} & \textbf{74.0} & \textbf{2{,}870 (349)} \\
    \bottomrule
  \end{tabular}%
  }
  \caption{\textbf{Per-benchmark and 4K/8K resolution breakdown on high-resolution development datasets} (Qwen3-VL-8B; V*Bench uses 3 seeds, HR-Bench evaluates 40 question units across 4K/8K and 4 option rotations over 2 seeds, and MME-RealWorld-Lite uses 2 seeds; Proc.\ Tok denotes shared-root processed visual and text tokens, with decoding tokens $C_{\mathrm{tok}}$ in parentheses). Context reset with Perceiver crops boosts V*Bench accuracy by $+10.9\%$ over 2-agent LatentMAS, preserves 8K accuracy ($76.3\%$ vs.\ $70.3\%$ without reset), and reduces shared-root processed tokens by $61\%$ ($2{,}870$ vs.\ $7{,}373$, $0.39\times$, at $1.04\times$ decoding tokens).}
  \label{tab:hires_breakdown}
\end{table*}

\paragraph{Oracle diagnostic scope and high-resolution search gains.}
\Cref{tab:hires_breakdown} and \cref{fig:supp-routing-highres}b expand \cref{tab:hires_clean}.
In the perception-error oracle diagnostic (12 selected development questions, 25 states, 3 samples per state), supplying the same verified visual fact yields $90.7\%$ accuracy in a reset context vs.\ $41.3\%$ when appended to the prior trace ($+49.3\%$, decoding ratio $1.10\times$, extra processed ratio $1.85\times$); note that fresh solves receive up to 2{,}048 new tokens whereas continuations receive $2{,}048 - k$ ($1{,}024$ after natural completion), so generation budget also contributes to this gap.
On high-resolution development benchmarks, pairing context reset with Perceiver crops reaches \textbf{89.2\%} on V*Bench ($+10.9\%$ over 2-agent global-image LatentMAS and $+12.5\%$ over random crops) and \textbf{55.7\%} on MME-RealWorld-Lite.

\paragraph{Resetting the context prevents 8K resolution degradation while cutting processed tokens by 61\%.}
When scaling from HR-Bench 4K to 8K (\cref{tab:hires_breakdown}), keeping the flawed trace (``No Reset + Perceiver Crops'') drops by $7.2\%$ ($77.5\% \to 70.3\%$) and increases decoding tokens to $599$ ($1.78\times$), whereas resetting the context holds 8K accuracy at \textbf{76.3\%} ($+6.0\%$) at $349$ decoding tokens ($1.04\times$).
Overall, Reset + Perceiver Crops raises macro accuracy to \textbf{74.0\%} while reducing shared-root processed tokens from $7{,}373$ to $2{,}870$ ($0.39\times$ vs.\ 2-agent LatentMAS and $0.36\times$ vs.\ Single VLM).

\subsection{Per-Benchmark Ablations and Text-Only Reproduction Details}
\label{app:textref}

\begin{table}[t]
  \centering\footnotesize
  \setlength{\tabcolsep}{3.5pt}
  \resizebox{\linewidth}{!}{%
  \begin{tabular}{@{}lccccc@{}}
    \toprule
    Ablation Variant (Qwen3-VL-8B) & MMStar & MathVista & WeMath & All & $C_{\mathrm{tok}}$ \\
    \midrule
    \rowcolor{ours}\textbf{\method{} (Full Two-Stage, Ours)} & \res{acc_ours_mmstar} & \res{acc_ours_mathvista} & \res{acc_ours_wemath} & \res{acc_ours} & \textbf{526} \\
    Stage 1 Early Stop Only (Solver Alone) & \res{acc_c1_mmstar} & \res{acc_c1_mathvista} & \res{acc_c1_wemath} & \res{acc_c1} & 564 \\
    Verify Every Run (No Trigger) & \textbf{\res{acc_always_mmstar}} & \res{acc_always_mathvista} & \textbf{\res{acc_always_wemath}} & \textbf{\res{acc_always}} & 937 \\
    w/o Local Crops (Global Image, Indep.) & \res{acc_si_mmstar} & \textbf{\res{acc_si_mathvista}} & \res{acc_si_wemath} & \res{acc_si} & 647 \\
    w/o Quick Check (Full Verifier Re-Solve) & 68.1 & 83.4 & 73.3 & 73.5 & 603 \\
    w/o Offline TRPS (Heuristic 1{,}024 tok) & 68.5 & 82.9 & 72.2 & 73.1 & 641 \\
    \bottomrule
  \end{tabular}%
  }
  \caption{\textbf{Per-benchmark accuracy and token breakdown of key ablation variants on Qwen3-VL-8B} across the 921-question test set. Full \method{} achieves the lowest decoding token cost ($526$ tokens), trading a modest gap on chart-heavy MathVista ($81.8\%$ vs.\ $83.6\%$ for independent-seed global-image verification at $647$ tokens).}
  \label{tab:ablation_sets}
\end{table}

\paragraph{Per-benchmark ablations highlight complementary domain behavior.}
In \cref{tab:ablation_sets}, full \method{} achieves the lowest decoding cost across all variants ($526$ tokens vs.\ $564$ for Stage~1 alone and $641$ for the 1{,}024-token heuristic).
The global-image control (which uses an independent Solver seed and disables Quick Check) attains higher point accuracy overall ($74.0\%$ vs.\ $73.5\%$; difference $-0.51\%$, 95\% CI $[-1.38, +0.43]$) and on chart-heavy MathVista ($83.6\%$ vs.\ $81.8\%$; $-1.82\%$, CI $[-3.65, -0.17]$), where tight crops can clip outer coordinate axes (\cref{app:limitations}), but consumes $23\%$ more decoding tokens ($647$ vs.\ $526$).

\begin{table}[t]
  \centering\footnotesize
  \setlength{\tabcolsep}{3pt}
  \resizebox{\linewidth}{!}{%
  \begin{tabular}{@{}lcccccc@{}}
    \toprule
    & \multicolumn{2}{c}{ARC-Easy} & \multicolumn{2}{c}{ARC-Challenge} & \multicolumn{2}{c}{GSM8K} \\
    \cmidrule(lr){2-3}\cmidrule(lr){4-5}\cmidrule(lr){6-7}
    Method (Qwen3-8B) & Paper~\cite{latentmas} & Ours & Paper~\cite{latentmas} & Ours & Paper~\cite{latentmas} & Ours \\
    \midrule
    Single LLM (Default) & \res{tr_single_arce} & \res{t_single_arce} & \res{tr_single_arcc} & \res{t_single_arcc} & \res{tr_single_gsm} & \res{t_single_gsm} \\
    TextMAS (4-Agent) & \res{tr_tm_arce} & \res{t_tm_arce} & \res{tr_tm_arcc} & \res{t_tm_arcc} & \res{tr_tm_gsm} & \res{t_tm_gsm} \\
    LatentMAS (4-Agent) & \res{tr_lmas_arce} & \res{t_lmas_arce} & \res{tr_lmas_arcc} & \res{t_lmas_arcc} & \res{tr_lmas_gsm} & \res{t_lmas_gsm} \\
    \rowcolor{ours}\textbf{\method{} (Stage 1 Only)} & -- & \textbf{\res{t_ours_arce}} & -- & \textbf{\res{t_ours_arcc}} & -- & \textbf{\res{t_ours_gsm}} \\
    \bottomrule
  \end{tabular}%
  }
  \caption{\textbf{Text benchmark reproduction on Qwen3-8B} across official test indices 300--599 (300 items per dataset, temperature $0.6$, 2{,}048-token cap) after fixing an upstream batched left-padding mask bug in LatentMAS~\cite{latentmas}. Operating via Stage~1 early stopping alone, \method{} outperforms bug-fixed LatentMAS on all three datasets with 20.2\% fewer tokens, while full 4-agent TextMAS reaches higher accuracy at $4.48\times$ token cost.}
  \label{tab:textref}
\end{table}

\paragraph{Fixing batched left-padding strengthens the text baseline.}
Reproducing Qwen3-8B results from LatentMAS~\cite{latentmas}, we fixed an upstream left-padding attention mask bug in batched inference (batch size $>1$), raising GSM8K development accuracy from $78.7\%$ to $91.7\%$.
\Cref{tab:textref} compares \cite{latentmas} with our matched 900-item evaluation (official test indices 300--599, 300 per dataset, temperature $0.6$, 2{,}048-token cap), where Stage~1 early stopping beats bug-fixed LatentMAS by $+1.9\%$ with $20.2\%$ fewer tokens (while Single+guard reaches $95.6\%$ at $741$ tokens and TextMAS reaches $96.2\%$ at $2{,}277$ tokens).

\subsection{Decoding Token Breakdown and B200 GPU Timing Details}
\label{app:matched-timing}

\begin{figure*}[t]
  \centering
  \includegraphics[width=0.95\textwidth]{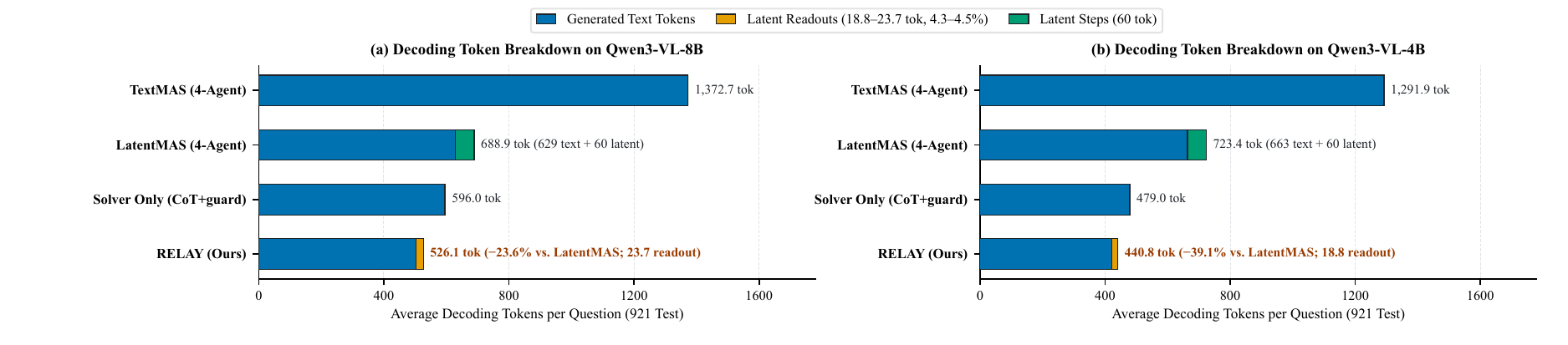}
  \vspace{-2mm}
  \caption{\textbf{Decoding token component breakdown on Qwen3-VL-8B (left) and Qwen3-VL-4B (right).}
    We decompose average decoding tokens per question across \res{n_confirm} test questions (3 seeds) into generated text tokens, latent readouts, and latent steps.
    Latent readouts add only $23.7$ tokens ($4.5\%$) on 8B and $18.8$ tokens ($4.3\%$) on 4B while eliminating 126--870 generated text tokens, reducing overall decoding token cost by 23.6\% on 8B and 39.1\% on 4B relative to 4-agent LatentMAS.}
  \label{fig:efficiency-components}
\end{figure*}

\begin{table}[t]
  \centering\footnotesize
  \setlength{\tabcolsep}{2.8pt}
  \resizebox{\linewidth}{!}{%
  \begin{tabular}{@{}llccccc@{}}
    \toprule
    Backbone & Method & MMStar & MathVista & WeMath & Overall (s/q) & Speedup \\
    \midrule
    \multirow{4}{*}{\textbf{8B}}
    & \textbf{\method{} (Ours)} & \textbf{8.23\,s} & \textbf{7.99\,s} & \textbf{21.20\,s} & \textbf{13.31\,s} & \textbf{1.00$\times$ (Ours)} \\
    & Solver Only (CoT+guard) & 6.23\,s & 9.02\,s & 25.63\,s & 14.52\,s & $1.09\times$ \\
    & LatentMAS (4-Agent) & 16.84\,s & 9.94\,s & 31.82\,s & 21.25\,s & \textbf{1.60$\times$ faster} \\
    & TextMAS (4-Agent) & 21.89\,s & 27.69\,s & 50.44\,s & 34.46\,s & \textbf{2.59$\times$ faster} \\
    \midrule
    \multirow{4}{*}{\textbf{4B}}
    & \textbf{\method{} (Ours)} & \textbf{7.91\,s} & \textbf{8.40\,s} & \textbf{19.51\,s} & \textbf{12.61\,s} & \textbf{1.00$\times$ (Ours)} \\
    & Solver Only (CoT+guard) & 8.08\,s & 5.54\,s & 18.98\,s & 11.84\,s & $0.94\times$ \\
    & LatentMAS (4-Agent) & 11.39\,s & 9.88\,s & 27.26\,s & 17.34\,s & \textbf{1.37$\times$ faster} \\
    & TextMAS (4-Agent) & 21.11\,s & 21.78\,s & 44.11\,s & 30.36\,s & \textbf{2.41$\times$ faster} \\
    \bottomrule
  \end{tabular}%
  }
  \caption{\textbf{Per-benchmark end-to-end B200 GPU wall-clock latency (seconds per question)} across 91 paired test questions at batch size~1 on Qwen3-VL-8B and Qwen3-VL-4B. Speedups are computed from full-precision means ($17.335/12.607 = 1.375 \to 1.37\times$ on 4B). \method{} accelerates fixed sequential multi-agent inference across all three benchmarks and both scales, while remaining within sampling variance of single-agent Solver Only (95\% latency-ratio CIs $[0.75, 1.10]$ on 8B and $[0.87, 1.29]$ on 4B).}
  \label{tab:matched-time}
\end{table}

\paragraph{Wall-clock timing protocol.}
We measure end-to-end latency on NVIDIA B200 GPUs at batch size~1 across 91 paired test questions (35 MMStar, 20 MathVista, and 36 WeMath, after single-item warm-up on two B200 workers) in counterbalanced order (\cref{tab:matched-time}).
The wall-clock timer covers initial image encoding $R$, latent readouts, Perceiver box prediction, crop encoding $C_{\mathrm{crop}}$, Quick Check, and Verifier decoding.
Note that these latency measurements reflect the 91-item standard-resolution paired subset rather than all 921 test items or high-resolution benchmarks.

\paragraph{Minimal readout overhead and multi-agent speedups.}
Across the 921-question test set (\cref{fig:efficiency-components}), latent readouts add only $23.7$ tokens ($4.5\%$) on 8B and $18.8$ tokens ($4.3\%$) on 4B while cutting generated text tokens from $628.9$ to $502.5$ on 8B and from $663.4$ to $421.9$ on 4B (and cutting 4B TextMAS from $1{,}291.9$ to $440.8$ total decoding tokens).
Consequently, \method{} accelerates end-to-end multi-agent inference by $1.60\times$--$2.59\times$ on 8B and $1.37\times$--$2.41\times$ on 4B across all three benchmarks.

\begin{figure*}[t]
  \centering
  \includegraphics[width=\textwidth]{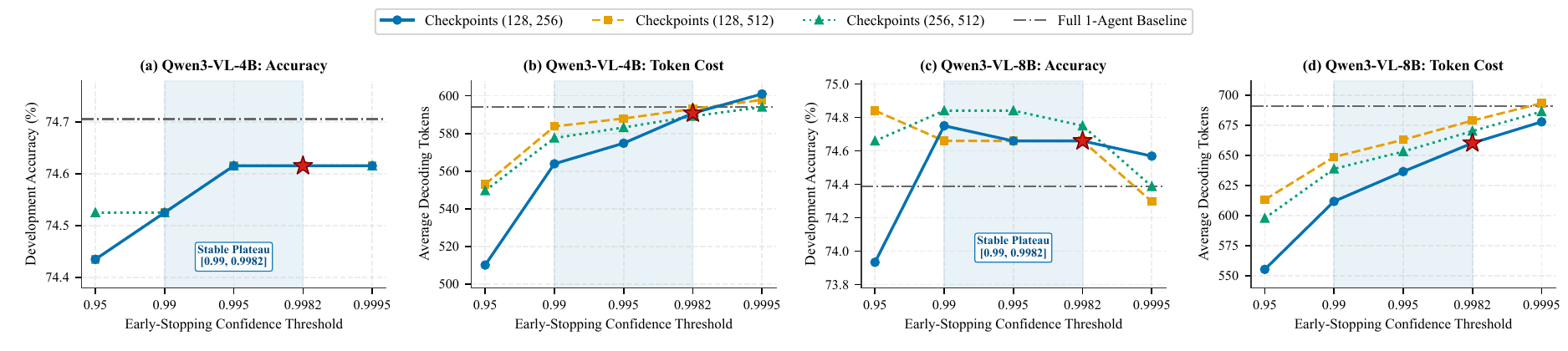}
  \vspace{-2mm}
  \caption{\textbf{Sensitivity of Stage 1 early stopping across checkpoint pairs and confidence thresholds.}
    We plot development accuracy and average decoding tokens on Qwen3-VL-4B (a, b) and Qwen3-VL-8B (c, d) across three early-stopping checkpoint pairs, namely $(128, 256)$, $(128, 512)$, and $(256, 512)$ tokens, and confidence thresholds from $0.95$ to $0.9995$.
    Thresholds between $0.99$ and $0.9982$ preserve or improve development accuracy across checkpoint pairs while cutting decoding tokens, with $(128, 256)$ yielding the largest savings.}
  \label{fig:settled-sensitivity}
\end{figure*}

\section{Ablation of Stage 1 and Stage 2 Stopping Signals}
\label{app:signal-components}

\begin{table*}[t]
\centering\footnotesize
\setlength{\tabcolsep}{4.5pt}
\resizebox{\textwidth}{!}{%
\begin{tabular}{@{}llcccc@{}}
\toprule
Model / Stage & Gating Variant & Accuracy (\%) & Decoding Tokens ($C_{\mathrm{tok}}$) & Acc.\ Diff.\ vs.\ Joint Gate & Token Cost vs.\ Joint Gate \\
\midrule
\multirow{4}{*}{Qwen3-VL-4B / Stage 1}
& No early stop (run to completion) & 74.71 & 594.1 & $0.00\%$ & $1.01\times$ \\
& Confidence only ($\tau_s = 0.9982$) & 74.71 & 579.3 & $0.00\%$ & $0.98\times$ \\
& Consistency only ($\tau_s = 0$) & 72.81 & 290.1 & $-1.90\%$ & $0.49\times$ \\
& \textbf{Consistency + Confidence (\method{})} & \textbf{74.71} & \textbf{590.7} & \textbf{Baseline} & \textbf{1.00$\times$} \\
\midrule
\multirow{4}{*}{Qwen3-VL-8B / Stage 1}
& No early stop (run to completion) & 74.39 & 690.8 & $-0.27\%$ & $1.05\times$ \\
& Confidence only ($\tau_s = 0.9982$) & 74.75 & 638.1 & $+0.09\%$ & $0.97\times$ \\
& Consistency only ($\tau_s = 0$) & 73.03 & 346.3 & $-1.64\%$ & $0.52\times$ \\
& \textbf{Consistency + Confidence (\method{})} & \textbf{74.66} & \textbf{660.2} & \textbf{Baseline} & \textbf{1.00$\times$} \\
\midrule
\multirow{4}{*}{Qwen3-VL-4B / Stage 2 Quick Check}
& No Quick Check (always full Verifier) & 75.16 & 618.4 & $-0.27\%$ & $1.12\times$ \\
& Confidence only ($\tau_g = 0.9$) & 75.16 & 539.1 & $-0.27\%$ & $0.98\times$ \\
& Consistency only ($\tau_g = 0$) & 75.43 & 512.9 & $0.00\%$ & $0.93\times$ \\
& \textbf{Consistency + Confidence (\method{})} & \textbf{75.43} & \textbf{552.2} & \textbf{Baseline} & \textbf{1.00$\times$} \\
\midrule
\multirow{4}{*}{Qwen3-VL-8B / Stage 2 Quick Check}
& No Quick Check (always full Verifier) & 75.48 & 765.8 & $+0.27\%$ & $1.15\times$ \\
& Confidence only ($\tau_g = 0.9$) & 75.20 & 620.8 & $0.00\%$ & $0.93\times$ \\
& Consistency only ($\tau_g = 0$) & 75.20 & 591.5 & $0.00\%$ & $0.89\times$ \\
& \textbf{Consistency + Confidence (\method{})} & \textbf{75.20} & \textbf{667.3} & \textbf{Baseline} & \textbf{1.00$\times$} \\
\bottomrule
\end{tabular}%
}
\caption{\textbf{Ablation of consistency and confidence signals in Stage 1 Early Stopping and Stage 2 Quick Check} on Qwen3-VL-4B (369 development items) and Qwen3-VL-8B (367 development items). Relying on answer consistency alone in Stage~1 drops accuracy by $1.64\%$--$1.90\%$, whereas confidence filtering prevents premature exits and Stage~2 Quick Check saves 12\%--15\% of tokens.}
\label{tab:signal-components}
\end{table*}

\begin{table}[t]
\centering\footnotesize
\renewcommand{\arraystretch}{0.92}
\setlength{\tabcolsep}{3pt}
\resizebox{\linewidth}{!}{%
\begin{tabular}{@{}llcccc@{}}
\toprule
Stage 1 Gating Variant & Benchmark & 4B Acc.\ (\%) & 4B Tokens & 8B Acc.\ (\%) & 8B Tokens \\
\midrule
\multirow{3}{*}{No early stop}
& MMStar & 70.59 & 450.1 & 72.08 & 444.9 \\
& MathVista & 84.00 & 572.6 & 82.67 & 650.4 \\
& WeMath & 71.78 & 722.6 & 70.67 & 909.5 \\
\midrule
\multirow{3}{*}{Confidence only}
& MMStar & 70.59 & 438.7 & 72.08 & 406.2 \\
& MathVista & 84.00 & 559.5 & 83.33 & 599.7 \\
& WeMath & 71.78 & 704.0 & 71.11 & 844.5 \\
\midrule
\multirow{3}{*}{Consistency only}
& MMStar & 68.35 & 265.3 & 70.37 & 292.9 \\
& MathVista & 77.00 & 277.3 & 79.33 & 305.1 \\
& WeMath & 73.56 & 318.4 & 70.89 & 415.4 \\
\midrule
\multirow{3}{*}{\textbf{Consistency + Confidence (\method{})}}
& MMStar & \textbf{70.59} & \textbf{448.7} & \textbf{72.08} & \textbf{417.4} \\
& MathVista & \textbf{84.00} & \textbf{569.8} & \textbf{83.33} & \textbf{613.6} \\
& WeMath & \textbf{71.78} & \textbf{717.3} & \textbf{70.89} & \textbf{880.7} \\
\bottomrule
\end{tabular}%
}
\caption{\textbf{Per-dataset breakdown of Stage 1 early-stopping signal variants} on Qwen3-VL-4B and Qwen3-VL-8B development sets. Without a confidence filter, consistency-only gating drops accuracy sharply on MathVista ($-7.0\%$ on 4B and $-4.0\%$ on 8B).}
\label{tab:signal-components-C1}
\end{table}

\paragraph{High confidence filtering prevents premature exits.}
On the 921-item test set (\cref{tab:ablation}), removing the confidence filter lowers Qwen3-VL-8B accuracy by $3.9\%$ ($73.5\% \to 69.6\%$), while removing consistency checking leaves accuracy virtually unchanged ($73.4\%$, $-0.04\%$, 95\% CI $[-0.19, +0.11]$) at $513$ tokens ($0.98\times$).
\Cref{fig:settled-sensitivity,tab:signal-components,tab:signal-components-C1} confirm this pattern on 4B and 8B development sets across $(128, 256)$, $(128, 512)$, and $(256, 512)$: consistency alone stops prematurely on uncertain guesses, losing up to $7.0\%$ on MathVista, whereas $\tau_s \in [0.99, 0.9982]$ prevents premature exits.

\newpage
\paragraph{One-step Quick Check avoids redundant Verifier decoding.}
In the lower half of \cref{tab:signal-components}, checking agreement between the Solver's 512-token answer and the Verifier's one-step readout ($\tau_g = 0.9$) on zoomed Perceiver crops cuts total decoding tokens by $12\%$--$15\%$ on development traces while keeping accuracy within $0.00\%$--$0.28\%$ of full Verifier reasoning (and cuts test tokens by $12.7\%$ with similar accuracy, 95\% CI $[-0.58, +0.64]$).
Quick Check thus confirms clear visual cases in one forward pass and reserves full Verifier generation for harder inputs.

\clearpage
\twocolumn[{%
\centering\footnotesize
\renewcommand{\arraystretch}{0.88}
\setlength{\tabcolsep}{5.2pt}
\resizebox{0.92\textwidth}{!}{%
\begin{tabular}{@{}ccccccccc@{}}
\toprule
& & & \multicolumn{3}{c}{\textbf{Qwen3-VL-4B ($N{=}369$)}} & \multicolumn{3}{c}{\textbf{Qwen3-VL-8B ($N{=}367$)}} \\
\cmidrule(lr){4-6}\cmidrule(lr){7-9}
Consistency Check & $\tau_s$ & $\tau_g$ & Acc.\ (\%) & $\Delta\text{Acc}$ vs.\ Solver & Cost Ratio & Acc.\ (\%) & $\Delta\text{Acc}$ vs.\ Solver & Cost Ratio \\
\midrule
Yes & 0.9500 & 0.9 & 74.89 & $+0.18\%$ & $0.81\times$ & 74.48 & $+0.09\%$ & $0.84\times$ \\
Yes & 0.9500 & 0.0 & 74.89 & $+0.18\%$ & $0.75\times$ & 74.39 & $0.00\%$ & $0.74\times$ \\
Yes & 0.9900 & 0.9 & 75.25 & $+0.54\%$ & $0.89\times$ & 75.11 & $+0.73\%$ & $0.92\times$ \\
Yes & 0.9900 & 0.0 & 75.25 & $+0.54\%$ & $0.83\times$ & 75.02 & $+0.64\%$ & $0.81\times$ \\
Yes & 0.9950 & 0.9 & 75.34 & $+0.63\%$ & $0.90\times$ & 75.02 & $+0.64\%$ & $0.94\times$ \\
Yes & 0.9950 & 0.0 & 75.34 & $+0.63\%$ & $0.84\times$ & 75.02 & $+0.64\%$ & $0.83\times$ \\
\rowcolor{ours}
\textbf{Yes (Locked Policy)} & \textbf{0.9982} & \textbf{0.9} & \textbf{75.43} & $\mathbf{+0.72\%}$ & $\mathbf{0.93\times}$ & \textbf{75.20} & $\mathbf{+0.82\%}$ & $\mathbf{0.97\times}$ \\
Yes & 0.9982 & 0.0 & 75.43 & $+0.72\%$ & $0.86\times$ & 75.20 & $+0.82\%$ & $0.86\times$ \\
\midrule
No & 0.9500 & 0.9 & 74.80 & $+0.09\%$ & $0.78\times$ & 74.21 & $-0.18\%$ & $0.78\times$ \\
No & 0.9500 & 0.0 & 74.80 & $+0.09\%$ & $0.72\times$ & 74.21 & $-0.18\%$ & $0.69\times$ \\
No & 0.9900 & 0.9 & 75.25 & $+0.54\%$ & $0.87\times$ & 75.20 & $+0.82\%$ & $0.87\times$ \\
No & 0.9900 & 0.0 & 75.25 & $+0.54\%$ & $0.80\times$ & 75.11 & $+0.73\%$ & $0.77\times$ \\
No & 0.9950 & 0.9 & 75.34 & $+0.63\%$ & $0.88\times$ & 74.93 & $+0.54\%$ & $0.91\times$ \\
No & 0.9950 & 0.0 & 75.34 & $+0.63\%$ & $0.82\times$ & 74.93 & $+0.54\%$ & $0.80\times$ \\
No & 0.9982 & 0.9 & 75.43 & $+0.72\%$ & $0.91\times$ & 75.11 & $+0.73\%$ & $0.94\times$ \\
No & 0.9982 & 0.0 & 75.43 & $+0.72\%$ & $0.84\times$ & 75.11 & $+0.73\%$ & $0.83\times$ \\
\bottomrule
\end{tabular}%
}
\captionsetup{hypcap=false}
\captionof{table}{\textbf{Complete 16-configuration threshold sweep via Offline TRPS on cached development sets} (Qwen3-VL-4B, $N{=}369$; Qwen3-VL-8B, $N{=}367$).
  We report pooled development accuracy, accuracy gain, and decoding cost ratio relative to full-length Solver Only across all 16 combinations of Stage~1 consistency checking, Stage~1 threshold $\tau_s$, and Stage~2 Quick Check threshold $\tau_g$ at $(128, 256, 512)$ tokens.
  Conservative thresholds ($\tau_s \in [0.99, 0.9982], \tau_g = 0.9$) form a broad plateau on development traces, with the locked policy achieving $75.43\%$ on 4B and $75.20\%$ on 8B.}
\label{tab:offline-grid-4b}
\vspace{4pt}
}]

\section{Offline TRPS Policy Selection and Threshold Sweep}
\label{app:offline-selector}

\begin{table}[t]
  \centering\footnotesize
  \setlength{\tabcolsep}{2.8pt}
  \resizebox{\linewidth}{!}{%
  \begin{tabular}{@{}lcccccc@{}}
    \toprule
    & & & \multicolumn{2}{c}{\textbf{Development Set}} & \multicolumn{2}{c}{\textbf{Frozen Test Set}} \\
    \cmidrule(lr){4-5}\cmidrule(lr){6-7}
    Candidate Policy & Horizon & Quick Check & Acc. & Cost & Acc. & Cost \\
    \midrule
    Stage 1 Only (No Relay) & -- & -- & 74.7\% & 0.81$\times$ & \res{fr_acc_c1}\% & 0.82$\times$ \\
    \textbf{\method{} (Selected)} & \textbf{512} & \textbf{\checkmark} & \textbf{75.2\%} & \textbf{0.82$\times$} & \textbf{\res{fr_acc_frozen}\%} & \textbf{0.76$\times$} \\
    Early Relay (Full Re-Solve) & 512 & -- & 75.5\% & 0.94$\times$ & \res{fr_acc_dev}\% & 0.88$\times$ \\
    Heuristic (w/o TRPS) & 1024 & -- & 74.6\% & 0.98$\times$ & 73.1\% & 0.93$\times$ \\
    Late Relay$^\dagger$ & 1024 & \checkmark & 76.2\% & 0.90$\times$ & \res{fr_acc_cut1024}\% & 0.85$\times$ \\
    Cap-Only Relay & 2048 & -- & 76.7\% & 1.09$\times$ & \res{fr_acc_cap}\% & 1.05$\times$ \\
    Verify Every Run & 0 & \checkmark & 75.8\% & 1.38$\times$ & \res{fr_acc_always}\% & 1.36$\times$ \\
    \bottomrule
  \end{tabular}%
  }
  \caption{\textbf{Offline TRPS structural policy comparison on the development set and frozen evaluation on the 921-item test set} (Qwen3-VL-8B; cost relative to 4-agent LatentMAS). Selection enforces the 95\% bootstrap upper bound $\widehat{U}_{0.95}(\rho_{\mathrm{tok}}) \le 0.95$ ($^\dagger$Late Relay has point cost $0.90\times$ but 95\% upper bound $0.986 > 0.95$, making it infeasible). Stage~1 Only has lower development cost ($0.81\times$), while the selected 512-token policy achieves higher development accuracy and the lowest test cost ($0.76\times$).}
  \label{tab:rsi_frontier}
  \vspace{-2mm}
\end{table}

\begin{table}[t]
  \centering\footnotesize
  \setlength{\tabcolsep}{3.0pt}
  \resizebox{\linewidth}{!}{%
  \begin{tabular}{@{}lcccc@{}}
    \toprule
    & \multicolumn{2}{c}{\textbf{Cross-Dataset \& Cross-Scale}} & \multicolumn{2}{c}{\textbf{Shared Dev Policy}} \\
    \cmidrule(lr){2-3}\cmidrule(lr){4-5}
    Target Backbone & $\Delta\text{Acc}$ [95\% CI] & Cost Ratio & $\Delta\text{Acc}$ [95\% CI] & Cost Ratio \\
    \midrule
    Qwen3-VL-4B & $0.00\%$ [$-1.78, +1.74$] & $\mathbf{0.75\times}$ ($-25\%$) & $+0.09\%$ [$-1.72, +1.90$] & $\mathbf{0.78\times}$ ($-22\%$) \\
    Qwen3-VL-8B & $-0.18\%$ [$-2.20, +1.89$] & $\mathbf{0.69\times}$ ($-31\%$) & $-0.18\%$ [$-2.39, +1.93$] & $\mathbf{0.78\times}$ ($-22\%$) \\
    \bottomrule
  \end{tabular}%
  }
  \caption{\textbf{Exploratory repeated-development replay of Offline TRPS} relative to cached Solver Only on development traces (369 items on 4B, 367 items on 8B). Thresholds tuned on two benchmarks of one model scale and replayed on the held-out third benchmark of the other scale cut decoder cost by 25\%--31\% ($0.75\times$ and $0.69\times$), while paired 95\% accuracy intervals remain wide and do not establish non-inferiority.}
  \label{tab:offline-selector-transfer}
  \vspace{-2mm}
\end{table}

\paragraph{Offline trace replay evaluates candidate policies in seconds.}
\Cref{tab:rsi_frontier} compares structural policies across relay horizons ($h \in \{0, 512, 1024, 2048\}$): under the 95\% upper-bound cost constraint $\widehat{U}_{0.95}(\rho_{\mathrm{tok}}) \le 0.95$ (\cref{eq:tradeoff_formulation}), Late Relay at $h=1024$ is infeasible (upper bound $0.986 > 0.95$), and TRPS selects $h=512$ with Perceiver crops and Quick Check, which improves development accuracy over Stage~1 ($74.7\% \to 75.2\%$) and achieves the lowest test token cost ($0.76\times$).
Fixing $(128, 256, 512)$, evaluating all 16 threshold combinations across 7 exploratory cross-dataset and cross-scale development tasks (112 scorings derived from 32 unique full replays, \cref{tab:offline-selector-transfer,tab:offline-grid-4b}) reads cached scalar readouts directly and completes the selector stage in \textbf{1.38 seconds} without extra VLM rollouts.

\paragraph{Exploratory repeated-development transfer scope.}
In \cref{tab:offline-grid-4b}, settings with $\tau_s \in \{0.99, 0.995, 0.9982\}$ improve pooled development accuracy by $+0.54\%$ to $+0.82\%$ over Solver Only while cutting cost to $0.77\times$--$0.97\times$ on both 4B and 8B, with our locked policy ($\text{Consistency}{=}\text{Yes}, \tau_s{=}0.9982, \tau_g{=}0.9$) reaching $75.43\%$ on 4B and $75.20\%$ on 8B.
Under the tighter cost-minimization objective used in exploratory leave-one-dataset/cross-scale replay (\cref{tab:offline-selector-transfer}), the selector picks the more aggressive $\text{Consistency}{=}\text{No}, \tau_s{=}0.95$ configuration in 6 of 7 tasks rather than the locked policy, reducing decoder cost by $25\%$--$31\%$ while leaving wide paired accuracy intervals ($[-1.78, +1.74]$ on 4B and $[-2.20, +1.89]$ on 8B).
Because this candidate family was previously informed by development traces, we treat \cref{tab:offline-selector-transfer} as exploratory replay rather than independent confirmation.

\clearpage
\twocolumn[{%
  \begin{center}
    \includegraphics[width=\textwidth]{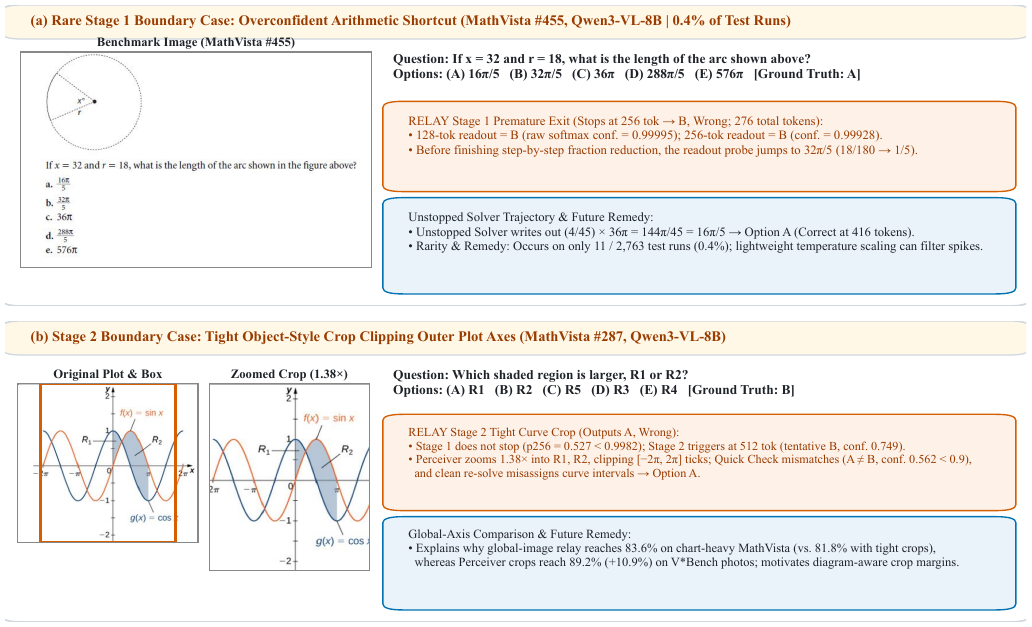}
    \vspace{-3mm}
    \captionsetup{hypcap=false}
    \captionof{figure}{\textbf{Representative test-set boundary cases and future extensions on MathVista (Qwen3-VL-8B).}
      \emph{(a) Rare Stage 1 overconfident arithmetic shortcut ($0.4\%$ of all test runs):} On MathVista \#455, an early mental simplification slip ($18/180 \to 1/5$) receives $>0.999$ raw softmax confidence at both 128 and 256 tokens, halting at 276 tokens before the unstopped Solver simplifies $144\pi/45 = 16\pi/5$ at 416 tokens.
      \emph{(b) Stage 2 tight crop clipping outer plot axes:} On MathVista \#287, Stage~1 does not stop ($p_{256}{=}0.527 < 0.9982$) and Stage~2 triggers at 512 tokens (tentative B, conf.\ $0.749$); zooming $1.38\times$ into shaded $R_1, R_2$ clips the outer $[-2\pi, 2\pi]$ ticks so Quick Check mismatches ($\text{A} \ne \text{B}$, conf.\ $0.562 < 0.9$) and clean re-solve outputs A, motivating diagram-aware crop margins.}
    \label{fig:failure-cases}
  \end{center}
  \vspace{4pt}
}]

\section{Qualitative Boundary Case Analyses}
\label{app:limitations}

Complementing the verified success cases in \cref{fig:case_study,fig:qual-cases}, \cref{fig:failure-cases} illustrates the two primary failure modes on the frozen 921-question test set.

\paragraph{Rare Stage 1 arithmetic shortcuts motivate lightweight confidence calibration.}
As noted in \cref{sec:conclusion} and \cref{fig:failure-cases}a (MathVista \#455), raw softmax confidence occasionally gives $>0.999$ probability to an early slip ($32\pi/5$ vs.\ $16\pi/5$) at 128 and 256 tokens before the unstopped Solver simplifies to $16\pi/5$ at 416 tokens.
Across 2{,}763 test runs, premature Stage~1 exits before a later self-correction occur on only 11 runs ($0.4\%$, 3 of 603 on MathVista).
Pairing readout agreement with temperature scaling or step-entropy smoothing provides a simple training-free safeguard.

\newpage
\paragraph{Tight crops on scientific plots motivate diagram-aware crop margins.}
\Cref{fig:failure-cases}b (MathVista \#287) illustrates the second boundary case in \cref{sec:conclusion}: Stage~1 correctly continues past 256 tokens ($p_{256}{=}0.527 < 0.9982$) and triggers Stage~2 at 512 tokens (tentative B, conf.\ $0.749$), but the $1.38\times$ Perceiver crop around the central curves clips the outer $[-2\pi, 2\pi]$ axis ticks.
On this clipped view, Quick Check disagrees ($\text{A} \ne \text{B}$, conf.\ $0.562 < 0.9$) and the clean Verifier misassigns the integration bounds to option A.
This explains why global-image verification reaches $83.6\%$ on chart-heavy MathVista vs.\ $81.8\%$ for tight local crops (\cref{tab:ablation_sets}), whereas Perceiver crops reach $89.2\%$ ($+10.9\%$) on natural photos in V*Bench (\cref{tab:hires_breakdown}).
Preserving full coordinate axes on charts while zooming tightly on natural images is a natural extension.

\end{document}